\documentclass[10pt,journal,compsoc,twocolumn]{IEEEtran}
\ifCLASSOPTIONcompsoc
  \usepackage[nocompress]{cite}
\else
  \usepackage{cite}
\fi

\ifCLASSINFOpdf
\else
\fi

\usepackage{graphicx}
\usepackage{amsmath,amssymb} 
\usepackage{booktabs}
\usepackage{enumerate}
\usepackage{color}
\usepackage[dvipsnames]{xcolor}
\usepackage{jeffe} 
\usepackage{subfigure}

\def\RedX{\color{red}{\sffamily X }}
\def\GreenY{\color{green!75!blue}\checkmark}

\usepackage[shortlabels]{enumitem} 
\setlist[itemize]{noitemsep, topsep=0pt}

\begin{document}
%

\title{Learning without Forgetting}
%
%
%
%

\author{Zhizhong~Li,
        Derek~Hoiem,~\IEEEmembership{Member,~IEEE}
\IEEEcompsocitemizethanks{\IEEEcompsocthanksitem Z. Li and D. Hoeim are with the Department
of Computer Science, University of Illinois, Urbana Champaign, IL, 61801.\protect\\
E-mail: \{zli115,dhoiem\}@illinois.edu}
}

%
%

\markboth{}%
{Shell \MakeLowercase{\textit{et al.}}: Bare Demo of IEEEtran.cls for Computer Society Journals}
%



\IEEEtitleabstractindextext{%
\begin{abstract}
When building a unified vision system or gradually adding new capabilities to a system, the usual assumption is that training data for all tasks is always available. However, as the number of tasks grows, storing and retraining on such data becomes infeasible. A new problem arises where we add new capabilities to a Convolutional Neural Network (CNN), but the training data for its existing capabilities are unavailable. We propose our Learning without Forgetting method, which uses only new task data to train the network while preserving the original capabilities.  Our method performs favorably compared to commonly used feature extraction and fine-tuning adaption techniques and performs similarly to multitask learning that uses original task data we assume unavailable. A more surprising observation is that Learning without Forgetting may be able to replace fine-tuning with similar old and new task datasets for improved new task performance.
\end{abstract}

}

\maketitle

\IEEEdisplaynontitleabstractindextext

%
\IEEEpeerreviewmaketitle

\IEEEraisesectionheading{\section{Introduction}\label{sec:introduction}}

%
%
%
%
\IEEEPARstart{M}{any} practical vision applications require learning new visual capabilities while maintaining performance on existing ones.  For example, a robot may be delivered to someone's house with a set of default object recognition capabilities, but new site-specific object models need to be added.  Or for construction safety, a system can identify whether a worker is wearing a safety vest or hard hat, but a superintendent may wish to add the ability to detect improper footware.  Ideally, the new tasks could be learned while sharing parameters from old ones, without suffering from Catastrophic Forgetting~\cite{mccloskey1989catastrophic,goodfellow2013empirical}  (degrading performance on old tasks) or having access to the old training data. Legacy data may be unrecorded, proprietary, or simply too cumbersome to use in training a new task.  This problem is similar in spirit to transfer, multitask, and lifelong learning. 

We aim at developing a simple but effective strategy on a variety of image classification problems with Convolutional Neural Network (CNN) classifiers. In our setting, a CNN has a set of shared parameters $\theta_s$ (e.g., five convolutional layers and two fully connected layers for AlexNet~\cite{krizhevsky2012imagenet} architecture), task-specific parameters for previously learned tasks $\theta_o$ (e.g., the output layer for ImageNet~\cite{ILSVRC15} classification and corresponding weights), and randomly initialized task-specific parameters for new tasks $\theta_n$ (e.g., scene classifiers).  It is useful to think of $\theta_o$ and $\theta_n$ as classifiers that operate on features parameterized by $\theta_s$.
Currently, there are three common approaches (Figures~\ref{fig:brief_comparison},~\ref{fig:methods_illust}) to learning $\theta_n$ while benefiting from previously learned $\theta_s$:

\begin {figure*}[t]
\centering
\caption{We wish to add new prediction tasks to an existing CNN vision system without requiring access to the training data for existing tasks. This table shows relative advantages of our method compared to commonly used methods.}
\begin{tabular}{@{\extracolsep{4pt}} r c c c c c @{} }
  \toprule
  & Fine & Duplicating and & Feature & Joint & Learning without \\
  & Tuning & Fine Tuning & Extraction & Training & Forgetting \\
  \midrule
new task performance	& good	& good	& \RedX medium	& best	& \GreenY best \\
original task performance	& \RedX bad	& good	& good	& good	& \GreenY good \\
training efficiency	& fast	& fast	& fast	& \RedX slow	& \GreenY fast \\
testing efficiency	& fast	& \RedX slow	& fast	& fast	& \GreenY fast \\
storage requirement	& medium	& \RedX large	& medium	& \RedX large	& \GreenY medium \\
requires previous task data	& no	& no	& no	& \RedX yes	& \GreenY no \\
  \bottomrule
\end{tabular}
\label{fig:brief_comparison}
\end{figure*}

\textbf{Feature Extraction} (e.g.,~\cite{Donahue_ICML2014DeCAF}): $\theta_s$ and $\theta_o$ are unchanged, and the outputs of one or more layers are used as features for the new task in training $\theta_n$.

\textbf{Fine-tuning} (e.g.,~\cite{Girshick_2014_CVPRRCNN}): $\theta_s$ and $\theta_n$ are optimized for the new task, while $\theta_o$ is fixed.  A low learning rate is typically used to prevent large drift in $\theta_s$. Potentially, the original network could be duplicated and fine-tuned for each new task to create a set of specialized networks.

It is also possible to use a variation of fine-tuning where part of $\theta_s$ -- the convolutional layers -- are frozen to prevent overfitting, and only top fully connected layers are fine-tuned.  This can be seen as a compromise between fine-tuning and feature extraction. In this work we call this method \textbf{Fine-tuning~FC} where FC stands for fully connected.

\textbf{Joint Training} (e.g.,~\cite{caruana1997multitask}): All parameters $\theta_s$, $\theta_o$, $\theta_n$ are jointly optimized, for example by interleaving samples from each task. This method's performance may be seen as an \textit{upper bound} of what our proposed method can achieve.

Each of these strategies has a major drawback. Feature extraction typically underperforms on the new task because the shared parameters fail to represent some information that is discriminative for the new task. 
Fine-tuning degrades performance on previously learned tasks because the shared parameters change without new guidance for the original task-specific prediction parameters. Duplicating and fine-tuning for each task results in linearly increasing test time as new tasks are added, rather than sharing computation for shared parameters. Fine-tuning~FC, as we show in our experiments, still degrades performance on the new task. Joint training becomes increasingly cumbersome in training as more tasks are learned and is not possible if the training data for previously learned tasks is unavailable.  

Besides these commonly used approaches, methods~\cite{furlanello2016active,jung2016less} have emerged that can continually add new prediction tasks by adapting shared parameters \textit{without access to training data} for previously learned tasks. (See Section~\ref{sec:related})

In this paper, we expand on our previous work~\cite{li2016learning}, \textbf{Learning without Forgetting} (LwF).  Using only examples for the new task, we optimize both for high accuracy for the new task and for preservation of responses on the existing tasks from the original network.   Our method is similar to joint training, except that our method does not need the old task's images and labels.  Clearly, if the network is preserved such that $\theta_o$ produces exactly the same outputs on all relevant images, the old task accuracy will be the same as the original network.  In practice, the images for the new task may provide a poor sampling of the original task domain, but our experiments show that preserving outputs on these examples is still an effective strategy to preserve performance on the old task and also has an unexpected benefit of acting as a regularizer to improve performance on the new task.  Our Learning without Forgetting approach has several advantages:
\begin{enumerate}[(1)]
\item Classification performance: Learning without Forgetting outperforms feature extraction and, more surprisingly, fine-tuning on the new task while greatly outperforming using fine-tuned parameters  $\theta_s$ on the old task.  Our method also generally perform better in experiments than recent alternatives~\cite{furlanello2016active,jung2016less}.
\item Computational efficiency: Training time is faster than joint training and only slightly slower than fine-tuning, and test time is faster than if one uses multiple fine-tuned networks for different tasks.
\item Simplicity in deployment: Once a task is learned, the training data does not need to be retained or reapplied to preserve performance in the adapting network.
\end{enumerate}

Compared to our previous work~\cite{li2016learning}, we conduct more extensive experiments. We compare to additional methods -- fine-tune FC, a commonly used baseline, and Less Forgetting Learning, a recently proposed method. We experiment on adjusting the balance between old-new task losses, providing a more thorough and intuitive comparison of related methods (Figure~\ref{fig:OvN}). We switch from the obsolete Places2 to a newer Places365-standard dataset. We perform stricter, more careful hyperparameter selection process, which slightly changed our results. We also include more detailed explanation of our method. Finally, we perform an experiment on application to video object tracking in Appendix~A.

\begin{figure*}[t]
  \centering
    \includegraphics[width=0.90\textwidth]{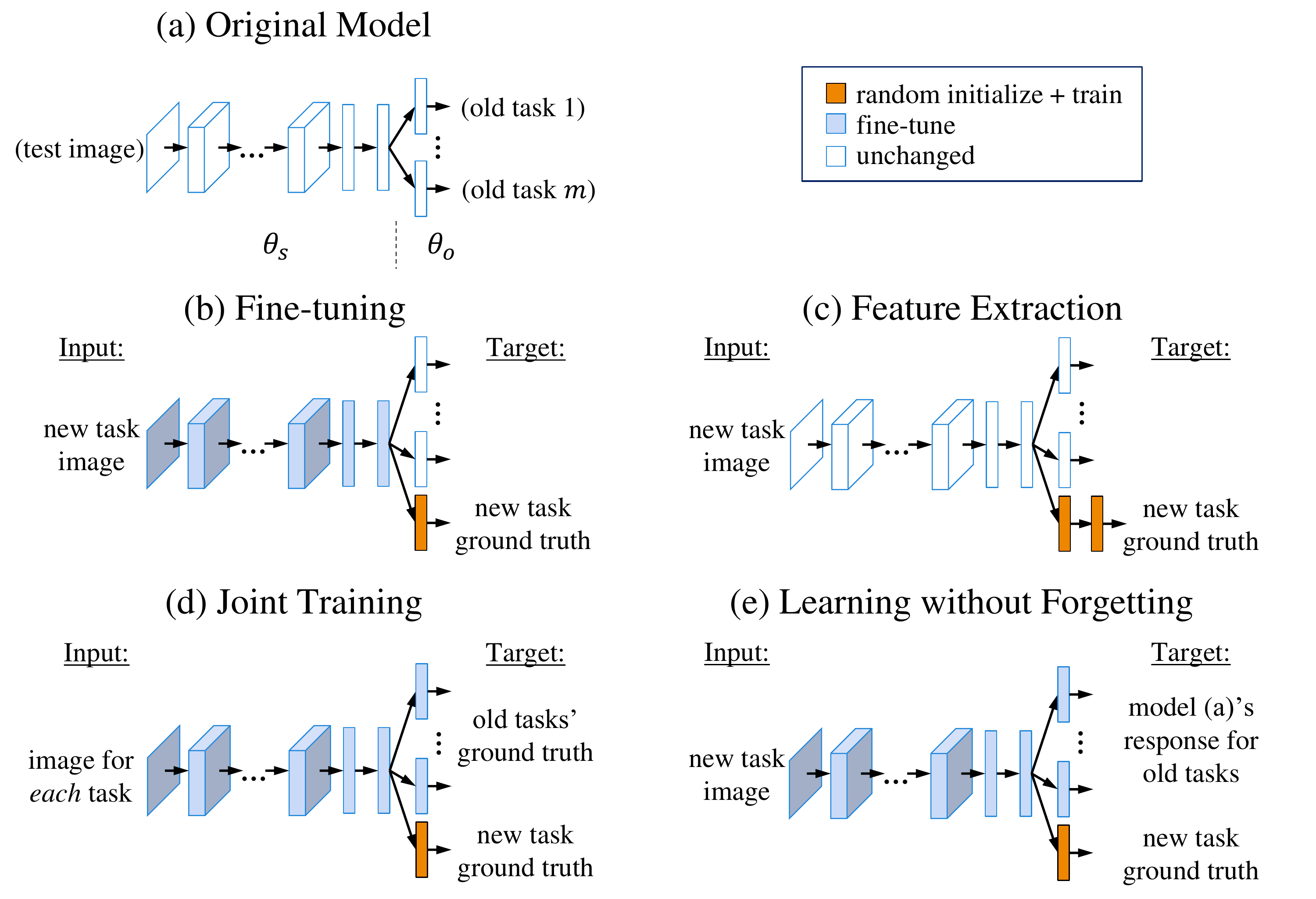}
    \caption{Illustration for our method (e) and methods we compare to (b-d). Images and labels used in training are shown. Data for different tasks are used in alternation in joint training.}
    \label{fig:methods_illust}
\end{figure*}

\section{Related work}
\label{sec:related}

Multi-task learning, transfer learning, and related methods have a long history.  In brief, our Learning without Forgetting approach could be seen as a combination of Distillation Networks~\cite{hinton2015distilling} and fine-tuning~\cite{Girshick_2014_CVPRRCNN}. Fine-tuning initializes with parameters from an existing network trained on a related data-rich problem and finds a new local minimum by optimizing parameters for a new task with a low learning rate.  The idea of Distillation Networks is to learn parameters in a simpler network that produce the same outputs as a more complex ensemble of networks either on the original training set or a large unlabeled set of data.  Our approach differs in that we solve for a set of parameters that works well on both old and new tasks using the same data to supervise learning of the new tasks and to provide unsupervised output guidance on the old tasks.

\subsection{Compared methods}

\textbf{Feature Extraction}~\cite{Donahue_ICML2014DeCAF,razavian2014cnnfeatures} uses a pre-trained deep CNN to compute features for an image. The extracted features are the activations of one layer (usually the last hidden layer) or multiple layers given the image. Classifiers trained on these features can achieve competitive results, sometimes outperforming human-engineered features~\cite{Donahue_ICML2014DeCAF}. Further studies~\cite{azizpour2014factors} show how hyper-parameters, e.g. original network structure, should be selected for better performance.  Feature extraction does not modify the original network and allows new tasks to benefit from complex features learned from previous tasks.  However, these features are not specialized for the new task and can often be improved by fine-tuning. 

\textbf{Fine-tuning}~\cite{Girshick_2014_CVPRRCNN} modifies the parameters of an existing CNN to train a new task. The output layer is extended with randomly intialized weights for the new task, and a small learning rate is used to tune all parameters from their original values to minimize the loss on the new task.  Sometimes, part of the network is frozen (e.g. the convolutional layers) to prevent overfitting. Using appropriate hyper-parameters for training, the resulting model often outperforms feature extraction~\cite{Girshick_2014_CVPRRCNN,azizpour2014factors} or learning from a randomly initialized network~\cite{agrawal14analyzing,yosinski2014howtransferable}.  Fine-tuning adapts the shared parameters $\theta_s$ to make them more discriminative for the new task, and the low learning rate is an indirect mechanism to preserve some of the representational structure learned in the original tasks.  Our method provides a more direct way to preserve representations that are important for the original task, improving both original and new task performance relative to fine-tuning in most experiments.


\textbf{Multitask learning} (e.g.,~\cite{caruana1997multitask}) aims to improve all tasks simultaneously by combining the common knowledge from all tasks. Each task provides extra training data for the parameters that are shared or constrained, serving as a form of regularization for the other tasks~\cite{chapelle2011boosted}.
For neural networks, Caruana~\cite{caruana1997multitask} gives a detailed study of multi-task learning. Usually the bottom layers of the network are shared, while the top layers are task-specific.
Multitask learning requires data from all tasks to be present, while our method requires only data for the new tasks.

\textbf{Adding new nodes} to each network layer is a way to preserve the original network parameters while learning new discriminative features. For example, Terekhov et al.~\cite{terekhov2015knowledge} propose Deep Block-Modular Neural Networks for fully-connected neural networks, and Rusu et al.~\cite{rusu2016progressive} propose Progressive Neural Networks for reinforcement learning.  Parameters for the original network are untouched, and newly added nodes are fully connected to the layer beneath them.  These methods has the downside of substantially expanding the number of parameters in the network, and can underperform~\cite{terekhov2015knowledge} both fine-tuning and feature extraction if insufficient training data is available to learn the new parameters, since they require a substantial number of parameters to be trained from scratch. We experiment with expanding the fully connected layers of original network but find that the expansion does not provide an improvement on our original approach.


\subsection{Topically relevant methods}

Our work also relates to \textbf{methods that transfer knowledge} between networks. Hinton et al.~\cite{hinton2015distilling} propose Knowledge Distillation, where knowledge is transferred from a large network or a network assembly to a smaller network for efficient deployment. The smaller network is trained using a modified cross-entropy loss (further described in Sec.~\ref{sec:LwF}) that encourages both large and small responses of the original and new network to be similar.  Romero et al.~\cite{romero2015fitnets} builds on this work to transfer to a deeper network by applying extra guidance on the middle layer. Chen et al.~\cite{Chen2016Net2Net} proposes the Net2Net method that immediately generates a deeper, wider network that is functionally equivalent to an existing one. This technique can quickly initialize networks for faster hyper-parameter exploration.  These methods aim to produce a differently structured network that approximates the original network, while we aim to find new parameters for the original network structure $(\theta_s, \theta_o)$ that approximate the original outputs while tuning shared parameters $\theta_s$ for new tasks.


Feature extraction and fine-tuning are special cases of \textbf{Domain Adaptation} (when old and new tasks are the same) or \textbf{Transfer Learning} (different tasks). These are different from multitask learning in that tasks are not simultaneously optimized. Transfer Learning uses knowledge from one task to help another, as surveyed by Pan et al.~\cite{pan2010survey}. 
The Deep Adaption Network by Long et al.~\cite{long2015learning} matches the RKHS embedding of the deep representation of both source and target tasks to reduce domain bias.  Another similar domain adaptation method is by Tzeng et al.~\cite{tzeng2015simultaneous}, which encourages the shared deep representation to be indistinguishable across domains. This method also uses knowledge distillation, but to help train the \textit{new} domain instead of preserving the old task.  Domain adaptation and transfer learning require that at least unlabeled data is present for both task domains.  In contrast, we are interested in the case when training data for the original tasks (i.e. source domains) are not available.

Methods that \textbf{integrate knowledge over time}, e.g. Lifelong Learning~\cite{thrun1998lifelong} and Never Ending Learning~\cite{NELL-aaai15}, are also related. Lifelong learning focuses on flexibly adding new tasks while transferring knowledge between tasks. Never Ending Learning focuses on building diverse knowledge and experience (e.g. by reading the web every day).  Though topically related to our work, these methods do not provide a way to preserve performance on existing tasks without the original training data.
Ruvolo et al.~\cite{eaton2013ella} describe a method to efficiently add new tasks to a multitask system, co-training all tasks while using only new task data. However, the method assumes that weights for all classifiers and regression models can be linearly decomposed into a set of bases.  In contrast with our method, the algorithm applies only to logistic or linear regression on engineered features, and these features cannot be made task-specific, e.g. by fine-tuning.

\subsection{Concurrently developed methods}

Concurrent with our previous work~\cite{li2016learning}, two methods have been proposed for continually add and integrate new tasks without using previous tasks' data.

\textbf{A-LTM}~\cite{furlanello2016active}, developed independently, is nearly identical in method but has very different experiments and conclusions. The main differences of method are in the weight decay regularization used for training and the warm-up step that we use prior to full fine-tuning.

However, we use large datasets to train our initial network (e.g. ImageNet) and then extend to new tasks from smaller datasets (e.g. PASCAL VOC), while A-LTM uses small datasets for the old task and large datasets for the new task.  The experiments in A-LTM~\cite{furlanello2016active} find much larger loss due to fine-tuning than we do, and the paper concludes that maintaining the data from the original task is necessary to maintain performance.  Our experiments, in contrast, show that we can maintain good performance for the old task while performing as well or sometimes better than fine-tuning for the new task, without access to original task data.  We believe the main difference is the choice of old-task new-task pairs and that we observe less of a drop in old-task performance from fine-tuning due to the choice (and in part to the warm-up step; see Table~\ref{tab:NOLOCK}).  We believe that our experiments, which start from a well-trained network and add tasks with less training data available, are better motivated from a practical perspective.

\textbf{Less Forgetting Learning}~\cite{jung2016less} is also a similar method, which preserves the old task performance by discouraging the shared representation to change. This method argues that the task-specific decision boundaries should not change, and keeps the old task's final layer unchanged, while our method discourages the old task output to change, and jointly optimizes both the shared representation and the final layer. We empirically show that our method outperforms Less Forgetting Learning on the new task.



\section{Learning without forgetting}

\label{sec:LwF}

\begin{figure*}[t]
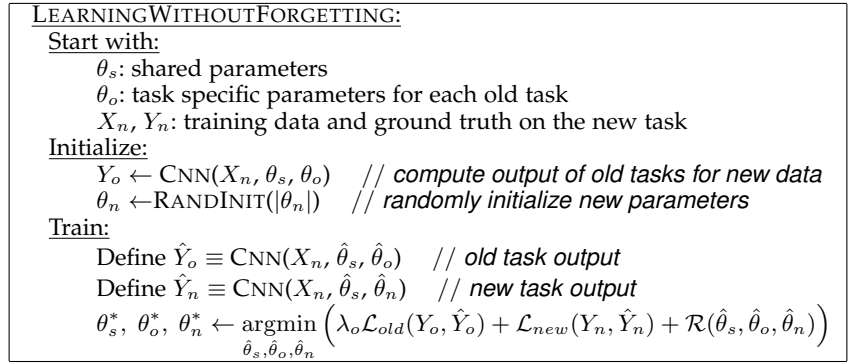

    \centering
    \begin{algo}
    	~\textul{\textsc{LearningWithoutForgetting}:}\+
    \\	\textul{Start with:}\+
    \\      $\theta_s$: shared parameters
    \\      $\theta_o$: task specific parameters for each old task
    \\      $X_n$, $Y_n$: training data and ground truth on the new task\-
    \\  \textul{Initialize:}\+
    \\  	$Y_o \gets$\textsc{ Cnn}($X_n$, $\theta_s$, $\theta_o$) \quad\Comment{compute output of old tasks for new data }
    \\      $\theta_n \gets$\textsc{RandInit}($|\theta_n|$) \quad\Comment{randomly initialize new parameters}\-
    \\  \textul{Train:}\+
    \\      Define $\hat{Y}_o \equiv$\textsc{ Cnn}($X_n$, $\hat{\theta}_s$, $\hat{\theta}_o$) \quad\Comment{old task output}
    \\      Define $\hat{Y}_n \equiv$\textsc{ Cnn}($X_n$, $\hat{\theta}_s$, $\hat{\theta}_n$) \quad\Comment{new task output}
    \\      $\displaystyle\theta_s^*,~\theta_o^*,~\theta_n^* \gets \argmin_{\hat\theta_s, \hat\theta_o, \hat\theta_n} \left( \lambda_o \mathcal{L}_{old}(Y_o,\hat{Y}_o) + \mathcal{L}_{new}(Y_n,\hat{Y}_n) + \mathcal{R}(\hat\theta_s, \hat\theta_o, \hat\theta_n) \right)$
    \end{algo}
    \caption{Procedure for Learning without Forgetting.}
    \label{fig:LwF_algorithm}
\end{figure*}

Given a CNN with shared parameters $\theta_s$ and task-specific parameters $\theta_o$ (Fig.~\ref{fig:methods_illust}(a)), our goal is to add task-specific parameters $\theta_n$ for a new task and to learn parameters that work well on old and new tasks, using images and labels from only the new task (i.e., \textit{without using data from existing tasks}).  Our algorithm is outlined in Fig.~\ref{fig:LwF_algorithm}, and the network structure illustrated in Fig.~\ref{fig:methods_illust}(e).

First, we record responses $\mathbf{y}_o$ on each new task image from the original network for outputs on the old tasks (defined by $\theta_s$ and $\theta_o$).  Our experiments involve classification, so the responses are the set of label probabilities for each training image.  Nodes for each new class are added to the output layer, fully connected to the layer beneath, with randomly initialized weights $\theta_n$.  The number of new parameters is equal to the number of new classes times the number of nodes in the last shared layer, typically a very small percent of the total number of parameters.  In our experiments (Sec.~\ref{sec:alternatives}), we also compare alternate ways of modifying the network for the new task. 

Next, we train the network to minimize loss for all tasks and regularization $\mathcal{R}$ using stochastic gradient descent.  The regularization $\mathcal{R}$ corresponds to a simple weight decay of 0.0005.  When training, we first freeze $\theta_s$ and $\theta_o$ and train $\theta_n$ to convergence (warm-up step).  Then, we jointly train all weights $\theta_s$, $\theta_o$, and $\theta_n$ until convergence (joint-optimize step). The warm-up step greatly enhances fine-tuning's old-task performance, but is not so crucial to either our method or the compared Less Forgetting Learning (see Table~\ref{tab:NOLOCK}). We still adopt this technique in Learning without Forgetting (as well as most compared methods) for the slight enhancement and a fair comparison.
 
For simplicity, we denote the loss functions, outputs, and ground truth for single examples.  The total loss is averaged over all images in a batch in training.  For new tasks, the loss encourages predictions $\mathbf{\hat{y}}_n$ to be consistent with the ground truth $\mathbf{y}_n$.  The tasks in our experiments are multiclass classification, so we use the common~\cite{krizhevsky2012imagenet,Simonyan14cVGG} multinomial logistic loss:
\begin{equation}
    \mathcal{L}_{new}(\mathbf{y}_n,\mathbf{\hat{y}}_n) = - \mathbf{y}_n \cdot \log\mathbf{\hat{y}}_n
    \label{eq:logistic_regression}
\end{equation}
where $\mathbf{\hat{y}}_n$ is the softmax output of the network and  $\mathbf{y}_n$ is the one-hot ground truth label vector.  If there are multiple new tasks, or if the task is multi-label classification where we make true/false predictions for each label, we take the sum of losses across the new tasks and the labels.


For each original task, we want the output probabilities for each image to be close to the recorded output from the original network.  We use the Knowledge Distillation loss, which was found by Hinton et al.~\cite{hinton2015distilling} to work well for encouraging the outputs of one network to approximate the outputs of another.
This is a modified cross-entropy loss that increases the weight for smaller probabilities:
\begin{align}
    \mathcal{L}_{old}(\mathbf{y}_o,\mathbf{\hat{y}}_o) &= - H(\mathbf{y}'_o,\mathbf{\hat{y}}'_o) \\
    &= - \sum_{i=1}^l y_o^{\prime (i)} \log \hat{y}_o^{\prime (i)}
\end{align}
where $l$ is the number of labels and $y_o^{\prime (i)}$, $\hat{y}_o^{\prime (i)}$ are the modified versions of recorded and current probabilities $y_o^{(i)}$, $\hat{y}_o^{(i)}$:
\begin{equation}
    y_o^{\prime (i)} = \dfrac{(y_o^{(i)})^{1/T}}{\sum_j (y_o^{(j)})^{1/T}}, \quad \hat{y}_o^{\prime (i)} = \dfrac{(\hat{y}_o^{(i)})^{1/T}}{\sum_j (\hat{y}_o^{(j)})^{1/T}}.
\end{equation}
If there are multiple old tasks, or if an old task is multi-label classification, we take the sum of the loss for each old task and label. Hinton et al.~\cite{hinton2015distilling} suggest that setting  $T>1$, which increases the weight of smaller logit values and encourages the network to better encode similarities among classes.
We use $T=2$ according to a grid search on a held out set, which aligns with the authors' recommendations.  In experiments, use of knowledge distillation loss leads to a slightly better but very similar performance to other reasonable losses. Therefore, it is important to constrain outputs for original tasks to be similar to the original network, but the similarity measure is not crucial.

$\lambda_o$ is a loss balance weight, set to 1 for most our experiments. Making $\lambda$ larger will favor the old task performance over the new task's, so we can obtain a old-task-new-task performance line by changing $\lambda_o$. (Figure~\ref{fig:OvN})

~\\\textbf{Relationship to joint training.}  As mentioned before, the main difference between joint training and our method is the need for the old dataset.  Joint training uses the old task's images and labels in training, while Learning without Forgetting no longer uses them, and instead uses the new task images $X_n$ and the recorded responses $Y_o$ as substitutes.  This eliminates the need to require and store the old dataset, brings us the benefit of joint optimization of the shared $\theta_s$, and also saves computation since the images $X_n$ only has to pass through the shared layers once for both the new task and the old task. However, the distribution of images from these tasks may be very different, and this substitution may potentially decrease performance. Therefore, joint training's performance may be seen as an upper-bound for our method.

~\\\textbf{Efficiency comparison.} The most computationally expensive part of using the neural network is evaluating or back-propagating through the shared parameters $\theta_s$, especially the convolutional layers. For training, feature extraction is the fastest because only the new task parameters are tuned.  LwF is slightly slower than fine-tuning because it needs to back-propagate through $\theta_o$ for old tasks but needs to evaluate and back-propagate through $\theta_s$ only once. Joint training is the slowest, because different images are used for different tasks, and each task requires separate back-propagation through the shared parameters.

All methods take approximately the same amount of time to evaluate a test image.  However, duplicating the network and fine-tuning for each task takes $m$ times as long to evaluate, where $m$ is the total number of tasks.

\subsection{Implementation details}
We use MatConvNet~\cite{vedaldi15matconvnet} to train our networks using stochastic gradient descent with momentum of 0.9 and dropout enabled in the fully connected layers. The data normalization (mean subtraction) of the original task is used for the new task. The resizing follows the implementation of the original network, which is $256\times 256$ for AlexNet and 256 pixels in the shortest edge with aspect ratio preserved for VGG. We randomly jitter the training data by taking random fixed-size crops of the resized images with offset on a $5\times 5$ grid, randomly mirroring the crop, and adding variance to the RGB values like in AlexNet~\cite{krizhevsky2012imagenet}. This data augmentation is applied to feature extraction too.

When training networks, we follow the standard practices for fine-tuning existing networks.  For random initialization of $\theta_n$, we use Xavier~\cite{glorot2010understanding} initialization. We use a learning rate much smaller than when training the original network ($0.1\sim0.02$ times the original rate). The learning rates are selected to maximize new task performance with a reasonable number of epochs.  For each scenario, the same learning rate are shared by all methods except feature extraction, which uses $5\times$ the learning rate due to its small number of parameters. 

We choose the number of epochs for both the warm-up step and the joint-optimize step based on validation on the held-out set.  We look at only the new task performance during validation. Therefore our selected hyperparameter favors the new task more. The compared methods converge at similar speeds, so we used the same number of epochs for each method for fair comparison; however, the convergence speed heavily depend on the original network and the task pair, and we validate for the number of epoch separately for each scenario. We perform stricter validation than in our previous work~\cite{li2016learning}, and the number of epochs is generally longer for each scenario. One exception is ImageNet$\rightarrow$Scene where we observe overfitting and have to shorten the training for feature extraction.  We lower the learning rate once by 10$\times$ at the epoch when the held out accuracy plateaus.  

To make a fair comparison, the intermediate network trained using our method (after the warm-up step) is used as a starting point for joint training and Fine Tuning, since this may speed up training convergence.  In other words, for each run of our experiment, we first freeze $\theta_s,\theta_o$ and train $\theta_n$, and use the resulting parameters to initialize our method, joint training and fine-tuning.  Feature extraction is trained separately because does not share the same network structure as our method.

For the feature extraction baseline, instead of extracting features at the last hidden layer of the original network (at the top of $\theta_s$), we freeze the shared parameters $\theta_s$, disable the dropout layers, and add a two-layer network with 4096 nodes in the hidden layer on top of it. This has the same effect of training a 2-layer network on the extracted features.  For joint training, loss for one task's output nodes is applied to only its own training images.  The same number of images are subsampled for every task in each epoch to balance their loss, and we interleave batches of different tasks for gradient descent.

\section{Experiments}

Our experiments are designed to evaluate whether Learning without Forgetting (LwF) is an effective method to learn a new task while preserving performance on old tasks.  We compare to common approaches of \textit{feature extraction}, \textit{fine-tuning}, and \textit{fine-tuning~FC}, and also \textit{Less Forgetting Learning} (LFL)~\cite{jung2016less}. These methods leverage an existing network for a new task without requiring training data for the original tasks.  Feature extraction maintains the exact performance on the original task.  We also compare to \textit{joint training} (sometimes called multitask learning) as an upper-bound on possible old task performance, since joint training uses images and labels for original and new tasks, while LwF uses only images and labels for the new tasks.

We experiment on a variety of image classification problems with varying degrees of inter-task similarity.  For the original (``old'') task, we consider the ILSVRC 2012 subset of \textit{ImageNet}~\cite{ILSVRC15} and the \textit{Places365-standard}~\cite{zhou2016places} dataset. Note that our previous work used Places2, a taster challenge in ILSVRC 2015~\cite{ILSVRC15} and an earlier version of Places365, but the dataset was deprecated after our publication.  ImageNet has 1,000 object category classes and more than 1,000,000 training images.  Places365 has 365 scene classes and $\sim 1,600,000$ training images.  We use these large datasets also because we assume we start from a well-trained network, which implies a large-scale dataset.  For the new tasks, we consider PASCAL \textit{VOC 2012 image classification}~\cite{Everingham15VOCretrospect} (``VOC''), \textit{Caltech-UCSD Birds-200-2011 fine-grained classification}~\cite{WahCUB_200_2011} (``CUB''), and \textit{MIT indoor scene classification}~\cite{quattoni2009recognizingindoor} (``Scenes'').  These datasets have a moderate number of images for training: 5,717 for VOC; 5,994 for CUB; and 5,360 for Scenes.  Among these, VOC is very similar to ImageNet, as subcategories of its labels can be found in ImageNet classes. MIT indoor scene dataset is in turn similar to Places365. CUB is dissimilar to both, since it includes only birds and requires capturing the fine details of the image to make a valid prediction. In one experiment, we use MNIST~\cite{lecun1998gradient} as the new task expecting our method to underperform, since the hand-written characters are completely unrelated to ImageNet classes.

We mainly use the AlexNet~\cite{krizhevsky2012imagenet} network structure because it is fast to train and well-studied by the community~\cite{yosinski2014howtransferable,azizpour2014factors,Girshick_2014_CVPRRCNN}. We also verify that similar results hold using 16-layer VGGnet~\cite{Simonyan14cVGG} on a smaller set of experiments.  For both network structures, the final layer (\verb|fc8|) is treated as task-specific, and the rest are shared ($\theta_s$) unless otherwise specified. The original networks pre-trained on ImageNet and Places365-standard are obtained from public online sources.

We report the center image crop mean average precision for VOC, and center image crop accuracy for all other tasks. We report the accuracy of the validation set of VOC, ImageNet and Places365, and on the test set of CUB and Scenes dataset. Since the test performance of the former three cannot be evaluated frequently, we only provide the performance on their test sets in one experiment. Due to the randomness within CNN training, we run our experiments three times, and report the mean performance.

Our experiments investigate adding a single new task to the network or adding multiple tasks one-by-one.  We also examine effect of dataset size and network design. In ablation studies, we examine alternative response-preserving losses, the utility of expanding the network structure, and fine-tuning with a lower learning rate as a method to preserve original task performance. Note that the results have multiple sources of variance, including random initialization and training, pre-determined termination (performance can fluctuate by training 1 or 2 additional epochs), etc.

\subsection{Main experiments}
\label{sec:mainexp}

\begin {table*}[t]
\centering
\caption{Performance for the single new task scenario.  For all tables, the \textbf{difference} of methods' performance with LwF (our method) is reported to facilitate comparison.  Mean Average Precision is reported for VOC and accuracy for all others. On the new task, LwF outperforms baselines in most scenarios, and performs comparably with joint training, which uses old task training data we consider unavailable for the other methods. On the old task, our method greatly outperforms fine-tuning and achieves slightly worse performance than joint training.  An exception is the ImageNet-MNIST task where LwF does not perform well on the old task.}
\subfigure[Using AlexNet structure (validation performance for ImageNet/Places365/VOC)]{
\resizebox{0.98\textwidth}{!}{
\begin{tabular}{@{\extracolsep{4pt}} r  c c c c c c c c c c c c c c @{} }
  \toprule
  & \multicolumn{2}{c}{ImageNet$\rightarrow$VOC}
  & \multicolumn{2}{c}{ImageNet$\rightarrow$CUB}
  & \multicolumn{2}{c}{ImageNet$\rightarrow$Scenes}
  & \multicolumn{2}{c}{Places365$\rightarrow$VOC}
  & \multicolumn{2}{c}{Places365$\rightarrow$CUB}
  & \multicolumn{2}{c}{Places365$\rightarrow$Scenes}
  & \multicolumn{2}{c}{ImageNet$\rightarrow$MNIST}
 \\ \cline{2-3} \cline{4-5} \cline{6-7} \cline{8-9} \cline{10-11} \cline{12-13} \cline{14-15} \\[-0.8em]
  & old & new & old & new & old & new & old & new & old & new & old & new & old & new \\
  \midrule
LwF (ours)	& 56.2	& 76.1	& 54.7	& 57.7	& 55.9	& 64.5	& 50.6	& 70.2	& 47.9	& 34.8	& 50.9	& 75.2	& 49.8	& 99.3 \\
\midrule
Fine-tuning	& -0.9	& -0.3	& -3.8	& -0.7	& -2.0	& -0.8	& -2.2	& 0.1	& -4.6	& 1.0	& -2.1	& -1.7	& -2.8	& 0.0 \\
LFL	& 0.0	& -0.4	& -1.9	& -2.6	& -0.3	& -0.9	& 0.2	& -0.7	& 0.7	& -1.7	& -0.2	& -0.5	& -2.9	& -0.6 \\
Fine-tune FC	& 0.5	& -0.7	& 0.2	& -3.9	& 0.6	& -2.1	& 0.5	& -1.3	& 1.8	& -4.9	& 0.3	& -1.1	& 7.0	& -0.2 \\
Feat. Extraction	& 0.8	& -0.5	& 2.3	& -5.2	& 1.2	& -3.3	& 1.1	& -1.4	& 3.8	& -12.3	& 0.8	& -1.7	& 7.3	& -0.8 \\
\midrule
Joint Training	& 0.7	& -0.2	& 0.6	& -1.1	& 0.5	& -0.6	& 0.7	& -0.0	& 2.3	& 1.5	& 0.3	& -0.3	& 7.2	& -0.0 \\
\bottomrule
\end{tabular}

\label{tab:6pairs}
}
}

\subfigure[Test set performance]{
\hspace{0.3cm}
\begin{tabular}{@{\extracolsep{4pt}} r  c c @{} }
  \toprule
  & \multicolumn{2}{c}{Places365$\rightarrow$VOC}
 \\ \cline{2-3} \\[-0.8em]
  & old & new \\
  \midrule
LwF (ours)	& 50.6	& 73.7 \\
\midrule
Fine-tuning	& -2.1	& 0.1 \\
Feat. Extraction	& 1.3	& -2.3 \\
\midrule
Joint Training	& 0.9	& -0.1 \\
\bottomrule
\end{tabular}

\label{tab:test}
\hspace{0.3cm}
}
\subfigure[Using VGGnet structure]{
\begin{tabular}{@{\extracolsep{4pt}} r  c c c c @{} }
  \toprule
  & \multicolumn{2}{c}{ImageNet$\rightarrow$CUB}
  & \multicolumn{2}{c}{ImageNet$\rightarrow$Scenes}
 \\ \cline{2-3} \cline{4-5} \\[-0.8em]
  & old & new & old & new \\
  \midrule
LwF (ours)	& 60.6	& 72.5	& 66.8	& 74.9 \\
\midrule
Fine-tuning	& -9.9	& 0.6	& -4.1	& -0.3 \\
LFL	& 0.3	& -2.8	& -0.0	& -2.1 \\
Fine-tune FC	& 3.2	& -6.7	& 1.4	& -2.4 \\
Feat. Extraction	& 8.2	& -8.6	& 1.9	& -5.1 \\
\midrule
Joint Training	& 8.0	& 2.5	& 4.1	& 1.5 \\
\bottomrule
\end{tabular}

\label{tab:VGG}

}
\end{table*}

\textbf{Single new task scenario.} First, we compare the results of learning one new task among different task pairs and different methods. Table~\ref{tab:6pairs},~\ref{tab:test} shows the performance of our method, and the relative performance of other methods compared to it using AlexNet. We also visualize the old-new performance comparison on two task pairs in Figure~\ref{fig:OvN}. We make the following observations:

\begin{itemize}
    \item[] \textit{On the new task, our method consistently outperforms fine-tuning, LFL, fine-tuning~FC, and feature extraction} except for ImageNet$\rightarrow$MNIST and Places365$\rightarrow$CUB using fine-tuning. The gain over fine-tuning was unexpected and indicates that preserving outputs on the old task is an effective regularizer.  (See Section~\ref{sec:discussion} for a brief discussion). This finding motivates replacing fine-tuning with LwF as the standard approach for adapting a network to a new task.
    \item[] \textit{On the old task, our method performs better than fine-tuning but often underperforms feature extraction, fine-tuning~FC, and sometimes LFL.} By changing shared parameters $\theta_s$, fine-tuning significantly degrades performance on the task for which the original network was trained.  By jointly adapting $\theta_s$ and $\theta_o$ to generate similar outputs to the original network on the old task, the performance loss is greatly reduced.
    \item[] \textit{Considering both tasks, Figure~\ref{fig:OvN} shows that if $\lambda_o$ is adjusted, LwF can perform better than LFL and fine-tuning~FC on the new task for the same old task performance} on the first task pair, and perform similarly to LFL on the second.  Indeed, fine-tuning~FC gives a performance between fine-tuning and feature extraction. LwF provides freedom of changing the shared representation compared to LFL, which may have boosted the new task performance.
    \item[] \textit{Our method performs similarly to joint training with AlexNet.} Our method tends to slightly outperform joint training on the new task but underperform on the old task, which we attribute to a different distribution in the two task datasets.  Overall, the methods perform similarly, a positive result since our method does not require access to the old task training data and is faster to train. Note that sometimes both tasks' performance degrade with $\lambda_o$ too large or too small. We suspect that making it too large essentially increases the old task learning rate, potentially making it suboptimal, and making it too small lessens the regularization.
    \item[] \textit{Dissimilar new tasks degrade old task performance more.}  For example, CUB is very dissimilar task from Places365~\cite{azizpour2014factors}, and adapting the network to CUB leads to a Places365 accuracy loss of $8.4\%$ ($3.8\%+4.6\%$) for fine-tuning, $3.8\%$ for LwF, and $1.5\%$ ($3.8\%-2.3\%$) for joint training.  In these cases, learning the new task causes considerable drift in the shared parameters, which cannot fully be accounted for by LwF because the distribution of CUB and Places365 images is very different.  Even joint training leads to more accuracy loss on the old task because it cannot find a set of shared parameters that works well for both tasks.  Our method does not outperform fine-tuning for Places365$\rightarrow$CUB and, as expected, ImageNet$\rightarrow$MNIST on the new task, since the hand-written characters provide poor indirect supervision for the old task. The old task accuracy drops substantially with fine-tuning and LwF, though more with fine-tuning.
    \item[] \textit{Similar observations hold for both VGG and AlexNet structures, except that joint training outperforms consistently for VGG, and LwF performs worse than before on the old task.} (Table~\ref{tab:VGG}) This indicates that these results are likely to hold for other network structures as well, though joint training may have a larger benefit on networks with more representational power. Among these results, LFL diverges using stochastic gradient descent, so we tuned down the learning rate ($0.5\times$) and used $\lambda_i=0.2$ instead.
\end{itemize}





\begin{figure*}[t]
  \centering
    \subfigure[Places365$\rightarrow$VOC]{
    \includegraphics[height=0.30\textheight]{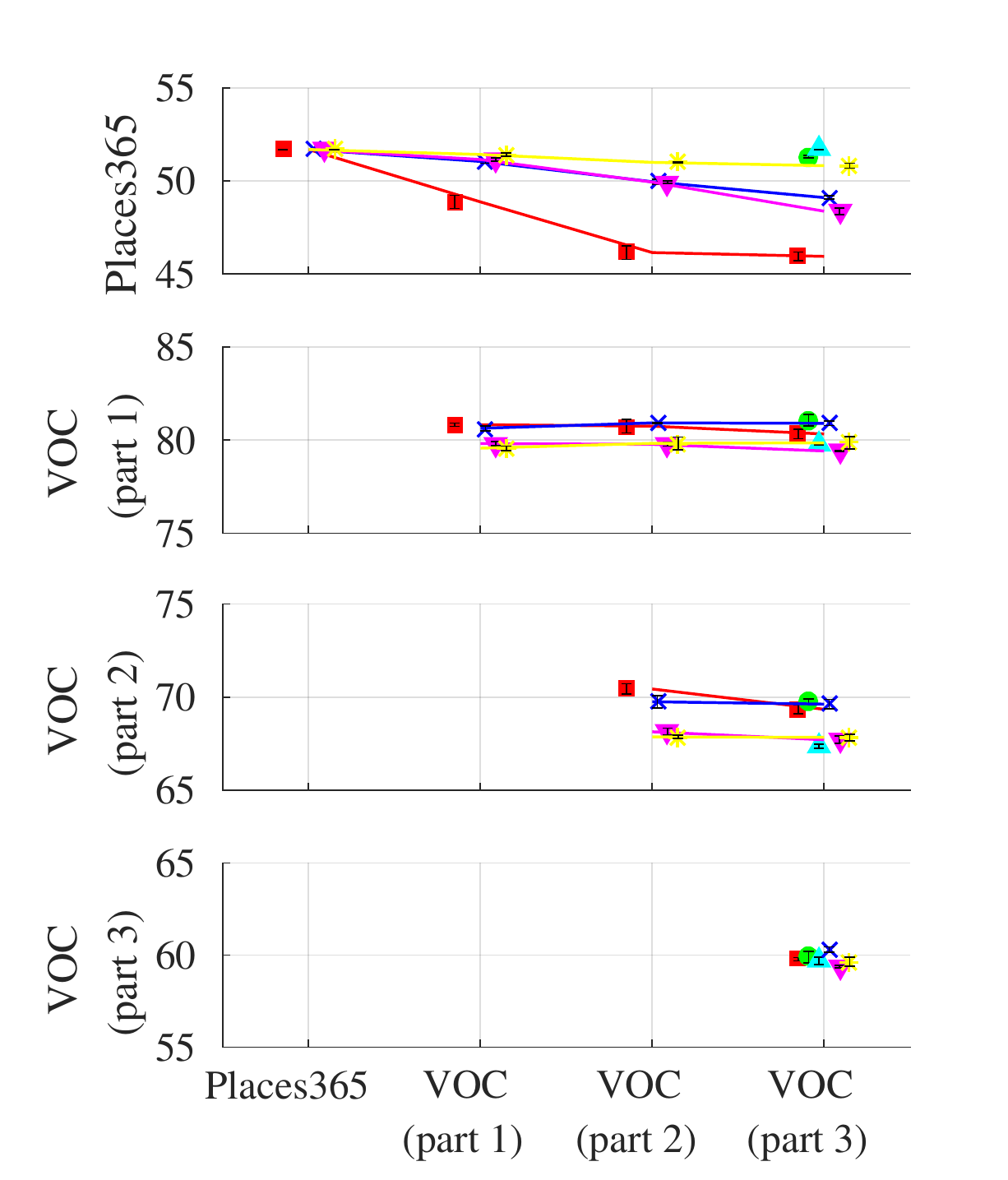}
    }
    \subfigure[ImageNet$\rightarrow$Scenes]{
    \includegraphics[height=0.30\textheight]{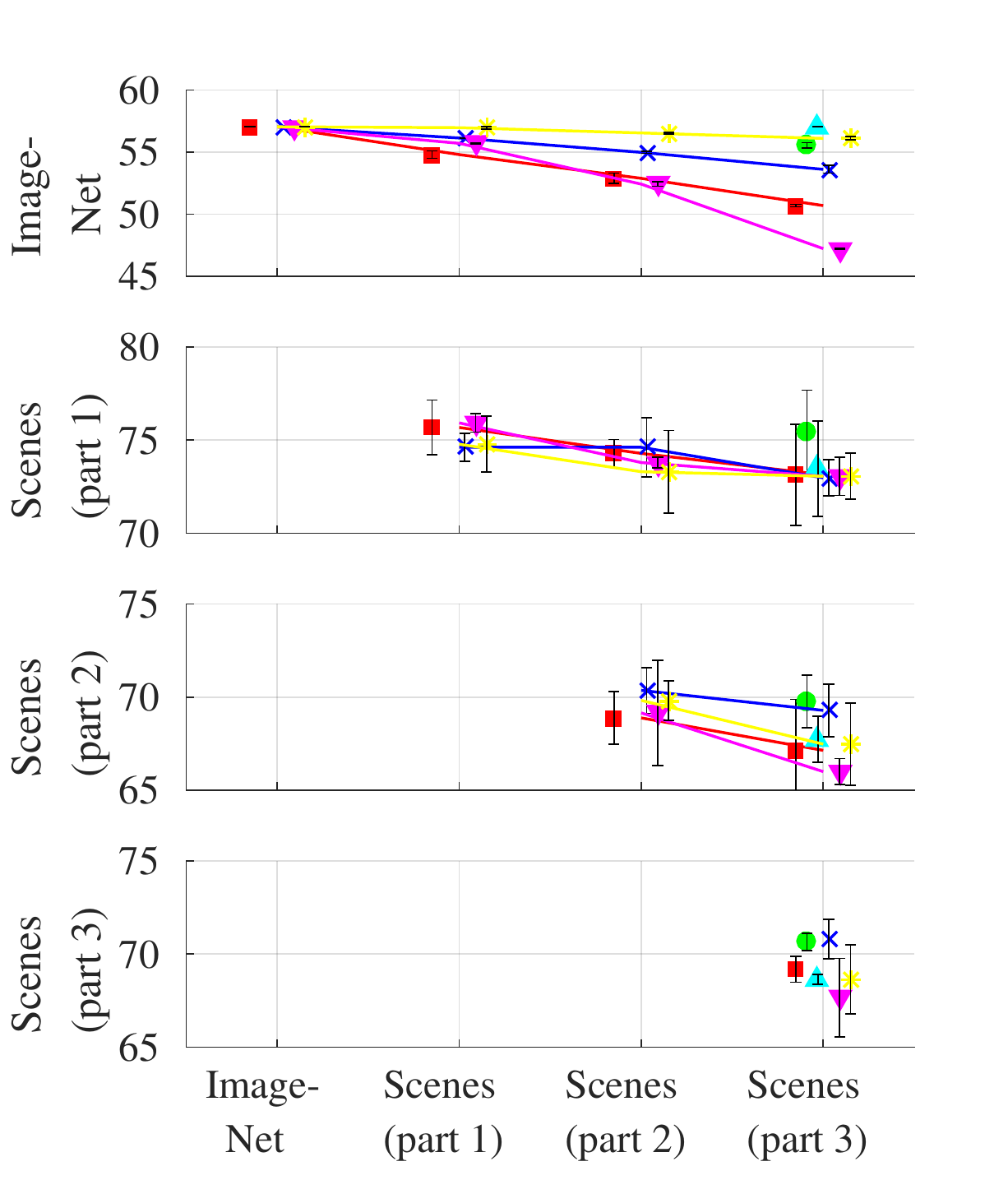}
    }
    \subfigure{
    \includegraphics[height=0.30\textheight]{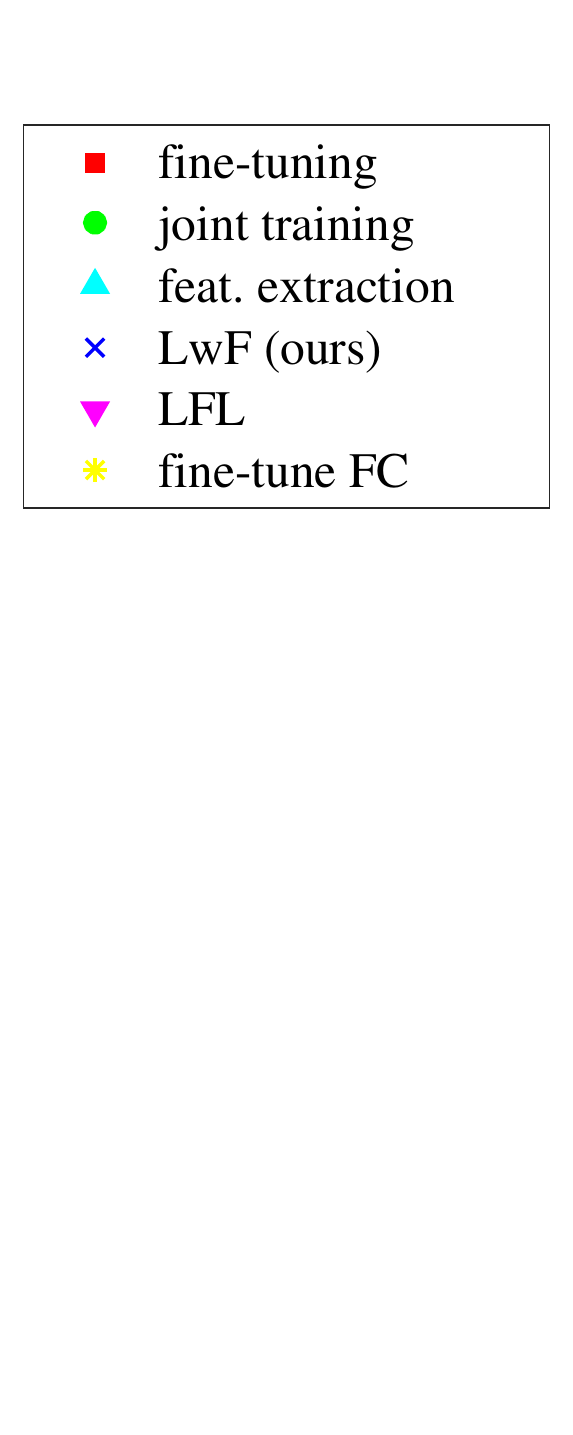}
    }
    \caption{Performance of each task when gradually adding new tasks to a pre-trained network. Different tasks are shown in different sub-graphs. The $x$-axis labels indicate the new task added to the network each time. Error bars shows $\pm 2$ standard deviations for 3 runs with different $\theta_n$ random initializations. Markers are jittered horizontally for visualization, but line plots are not jittered to facilitate comparison. For all tasks, our method degrades slower over time than fine-tuning and outperforms feature extraction in most scenarios. For Places2$\rightarrow$VOC, our method performs comparably to joint training. }
    \label{fig:accumulative}
\end{figure*}

\noindent\textbf{Multiple new task scenario.} Second, we compare different methods when we cumulatively add new tasks to the system, simulating a scenario in which new object or scene categories are gradually added to the prediction vocabulary. We experiment on gradually adding VOC task to AlexNet trained on Places365, and adding Scene task to AlexNet trained on ImageNet. These pairs have moderate difference between original task and new tasks. We split the new task classes into three parts according to their similarity -- VOC into transport, animals and objects, and Scenes into large rooms, medium rooms and small rooms. (See supplemental material) The images in Scenes are split into these three subsets.  Since VOC is a multilabel dataset, it is not possible to split the images into different categories, so the labels are split for each task and images are shared among all the tasks.  

Each time a new task is added, the responses of all other tasks $Y_o$ are re-computed, to emulate the situation where data for \textit{all} original tasks are unavailable. Therefore, $Y_o$ for older tasks changes each time. For feature extractor and joint training, cumulative training does not apply, so we only report their performance on the final stage where all tasks are added. Figure~\ref{fig:accumulative} shows the results on both dataset pairs. Our findings are usually consistent with the single new task experiment: \textit{LwF outperforms fine-tuning, feature extraction, LFL, and fine-tuning~FC for most newly added tasks. However, LwF performs similarly to joint training only on newly added tasks (except for Scenes part 1), and underperforms joint training on the old task after more tasks are added.}


\begin{figure*}[t]
  \centering
    \subfigure[CUB accuracy (new)]{
    \includegraphics[height=0.22\textheight]{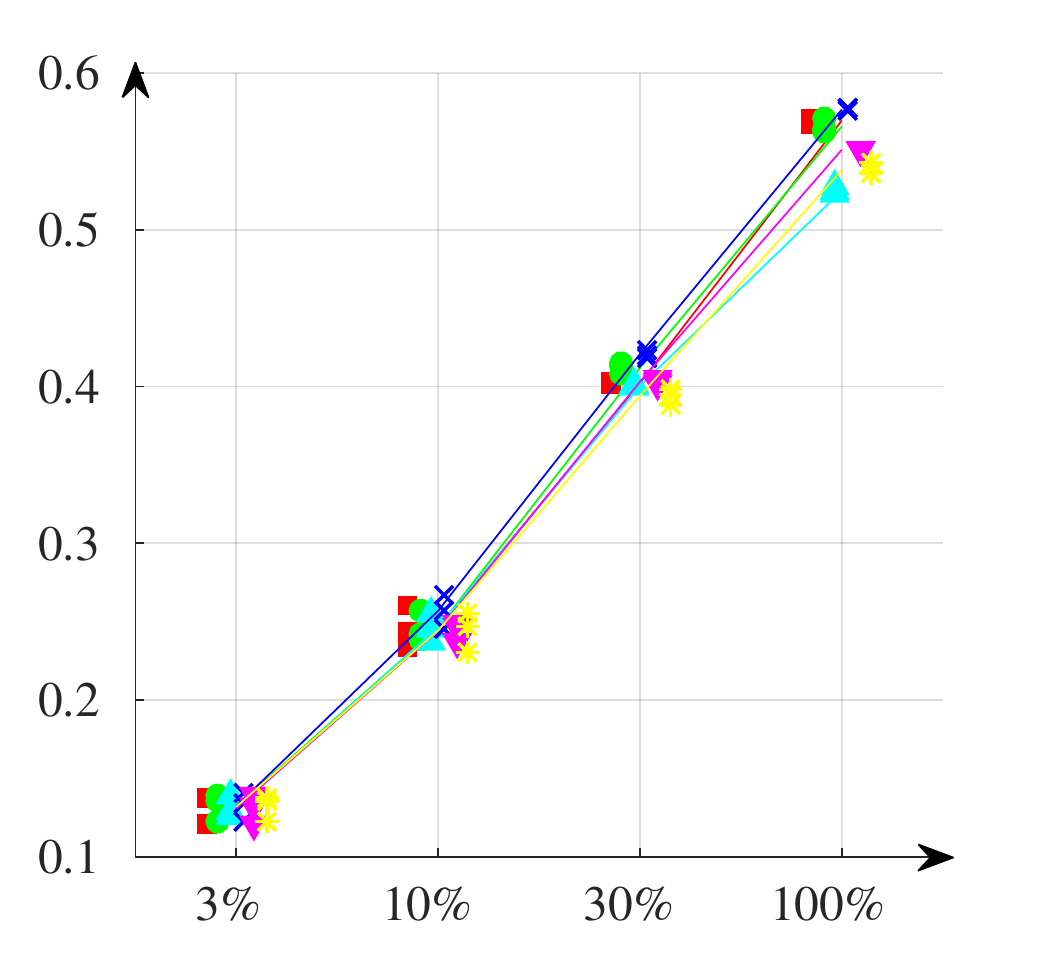}
    }
    \subfigure[ImageNet accuracy (old)]{
    \hspace{1em}
    \includegraphics[height=0.22\textheight]{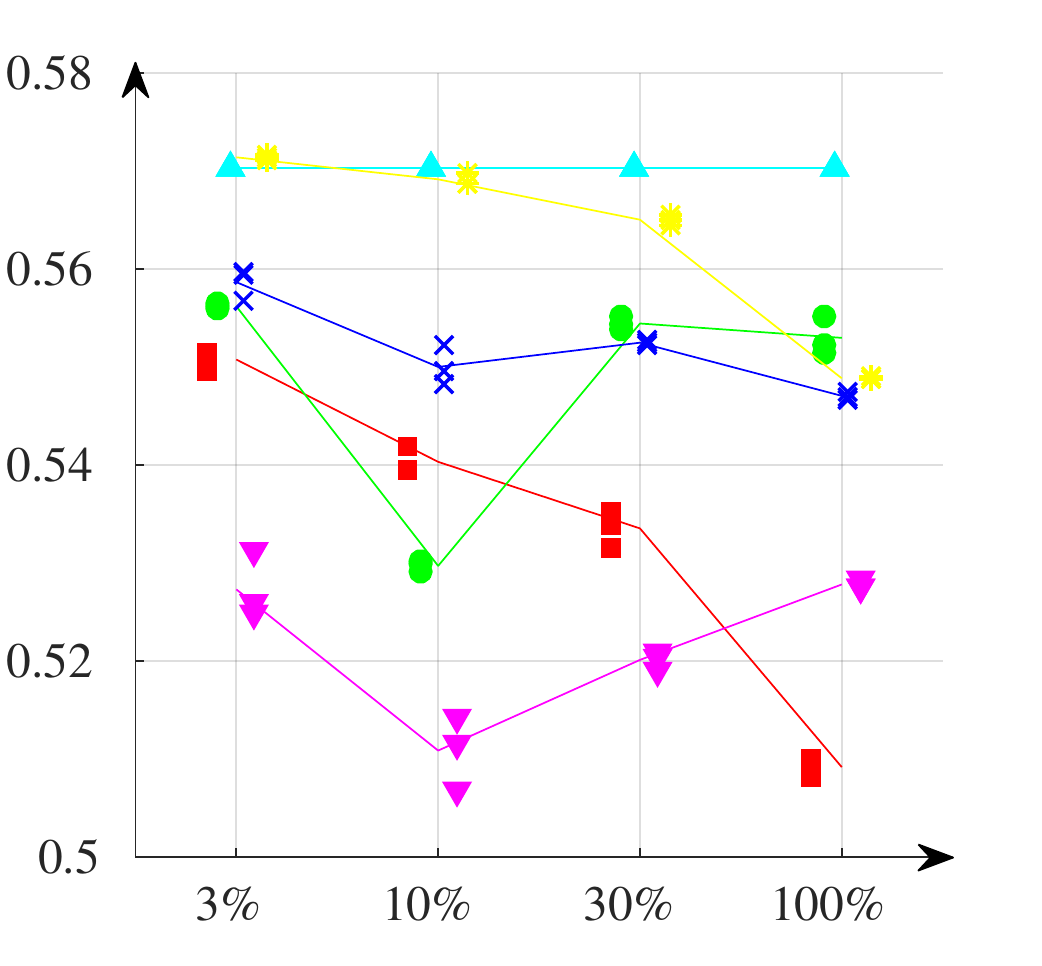}
    \hspace{1em}
    }
    \subfigure{
    \includegraphics[height=0.22\textheight]{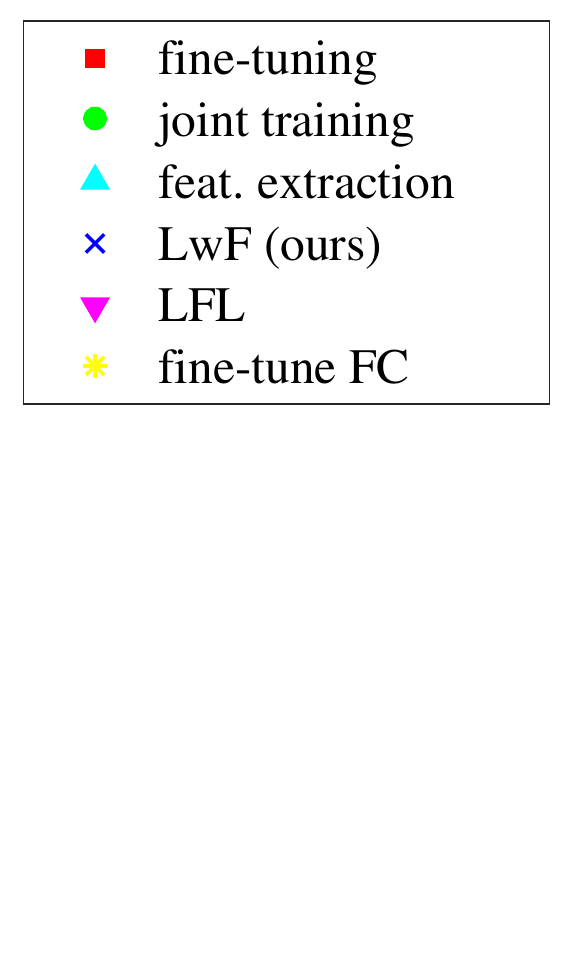}
    }
    \caption{Influence of subsampling new task training set on compared methods. The $x$-axis indicates diminishing training set size. Three runs of our experiments with different random $\theta_n$ initialization and dataset subsampling are shown. Scatter points are jittered horizontally for visualization, but line plots are not jittered to facilitate comparison. Differences between LwF and compared methods on both the old task and the new task decrease with less data, but the observations remain the same. LwF outperforms fine-tuning despite the change in training set size.}
    \label{fig:TRDATA}
\end{figure*}

~\\\textbf{Influence of dataset size.} We inspect whether the size of the new task dataset affects our performance relative to other methods. We perform this experiment on adding CUB to ImageNet AlexNet. We subsample the CUB dataset to 30\%, 10\% and 3\% when training the network, and report the result on the entire validation set. Note that for joint training, since each dataset has a different size, the same number of images are subsampled to train both tasks (resampled each epoch), which means a smaller number of ImageNet images being used at one time. Our results are shown in Figure~\ref{fig:TRDATA}. Results show that \textit{the same observations hold.  Our method outperforms fine-tuning on both tasks.  Differences between methods tend to increase with more data used, although the correlation is not definitive.}

\begin{figure*}[t]
  \centering
    \includegraphics[width=0.90\textwidth]{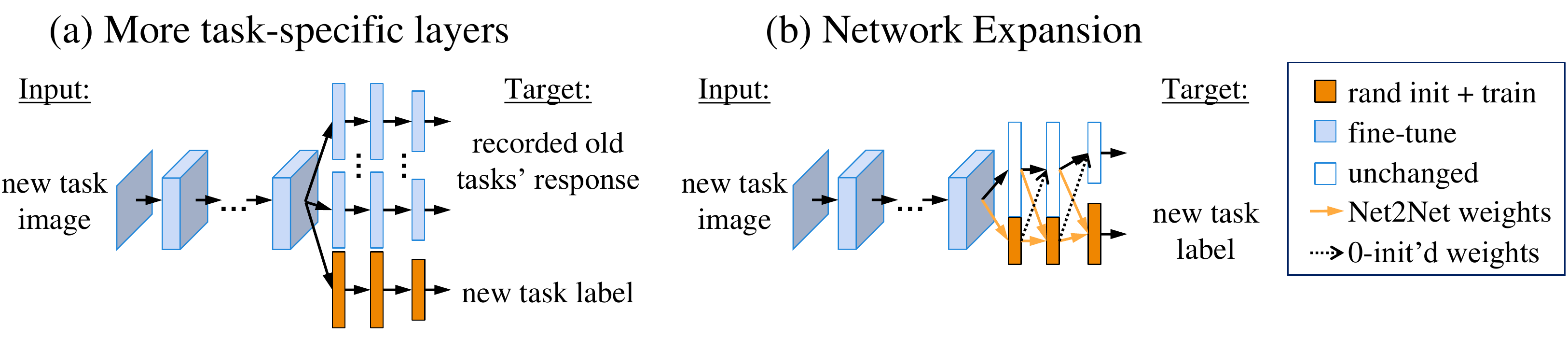}
    \caption{Illustration for alternative network modification methods.  In (a), more fully connected layers are task-specific, rather than shared.  In (b), nodes for multiple old tasks (not shown) are connected in the same way.  LwF can also be applied to Network Expansion by unfreezing all nodes and matching output responses on the old tasks.}
    \label{fig:methods_illust_alt}
\end{figure*}

\begin{table*}[t]
    \centering
    \caption{Performance of our method versus various alternative design choices. In most cases, these alternative choices do not provide consistent advantage or disadvantage compared to our method.}
    \subfigure[Changing the number of task-specific layers, using network expansion, or attempting to lower $\theta_s$'s learning rate when fine-tuning. ]{
\begin{tabular}{@{\extracolsep{4pt}} r  c c c c c c @{} }
  \toprule
  & \multicolumn{2}{c}{ImageNet$\rightarrow$CUB}
  & \multicolumn{2}{c}{ImageNet$\rightarrow$Scenes}
  & \multicolumn{2}{c}{Places365$\rightarrow$VOC}
 \\ \cline{2-3} \cline{4-5} \cline{6-7} \\[-0.8em]
  & old & new & old & new & old & new \\
  \midrule
LwF at output layer (ours)	& 54.7	& 57.7	& 55.9	& 64.5	& 50.6	& 70.2 \\
last hidden layer	& 54.7	& 56.2	& 55.7	& 65.0	& 50.7	& 70.6 \\
$2^{nd}$ last hidden (Fig. 6(a))	& 54.6	& 57.1	& 55.8	& 64.2	& 50.8	& 70.5 \\
\midrule
network expansion	& 57.0	& 54.0	& 57.0	& 62.5	& 51.7	& 67.1 \\
network expansion + LwF	& 54.4	& 57.0	& 55.7	& 63.9	& 50.7	& 70.4 \\
\midrule
fine-tuning (10\% $\theta_s$ learning rate)	& 52.2	& 54.9	& 54.8	& 62.7	& 49.3	& 69.5 \\
\bottomrule
\end{tabular}

     \label{tab:BRANCH_EXPAND_LOWLR} }
    
    \subfigure[Performing LwF and fine-tuning with and without warmup. The warmup step is not crucial for LwF, but is essential for fine-tuning's old task performance. ]{
\begin{tabular}{@{\extracolsep{4pt}} r  c c c c c c @{} }
  \toprule
  & \multicolumn{2}{c}{ImageNet$\rightarrow$CUB}
  & \multicolumn{2}{c}{ImageNet$\rightarrow$Scenes}
  & \multicolumn{2}{c}{Places365$\rightarrow$VOC}
 \\ \cline{2-3} \cline{4-5} \cline{6-7} \\[-0.8em]
  & old & new & old & new & old & new \\
  \midrule
LwF	& 54.7	& 57.7	& 55.9	& 64.5	& 50.6	& 70.2 \\
fine-tuning	& 50.9	& 57.0	& 53.9	& 63.8	& 48.4	& 70.3 \\
LFL	& 52.8	& 55.1	& 55.5	& 63.6	& 50.8	& 69.5 \\
\midrule
LwF (no warm-up)	& 53.5	& 59.9	& 55.2	& 64.9	& 50.4	& 70.0 \\
fine-tuning (no warm-up)	& 42.5	& 59.8	& 49.8	& 63.9	& 42.3	& 70.0 \\
LFL (no warm-up)	& 52.5	& 55.3	& 55.4	& 63.0	& 50.6	& 69.1 \\
\bottomrule
\end{tabular}

     \label{tab:NOLOCK} }
    \label{tab:alternatives}
\end{table*}

\begin{figure*}[t]
  \centering
    \subfigure[Places365$\rightarrow$VOC]{
    \includegraphics[height=0.22\textheight]{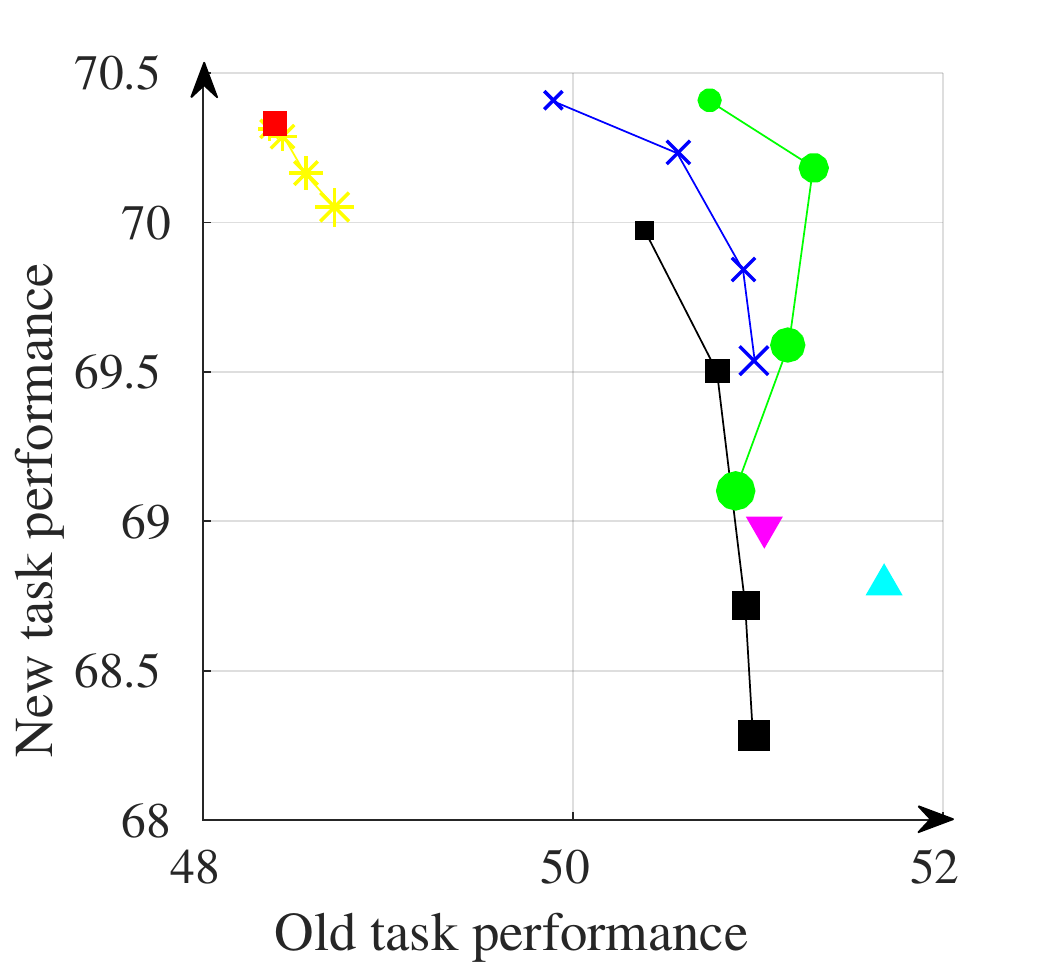}
    }
    \subfigure[ImageNet$\rightarrow$Scene]{
    \hspace{1em}
    \includegraphics[height=0.22\textheight]{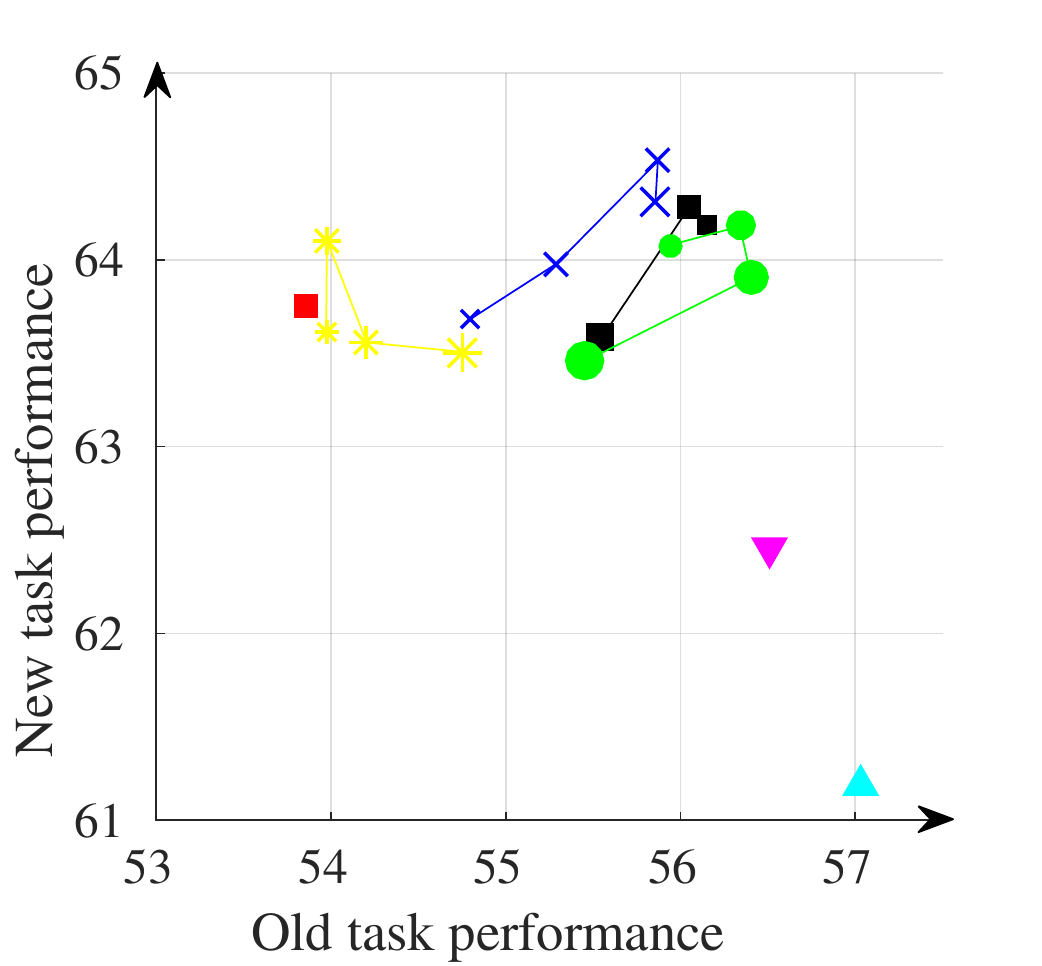}
    \hspace{1em}
    }
    \subfigure{
    \includegraphics[height=0.22\textheight]{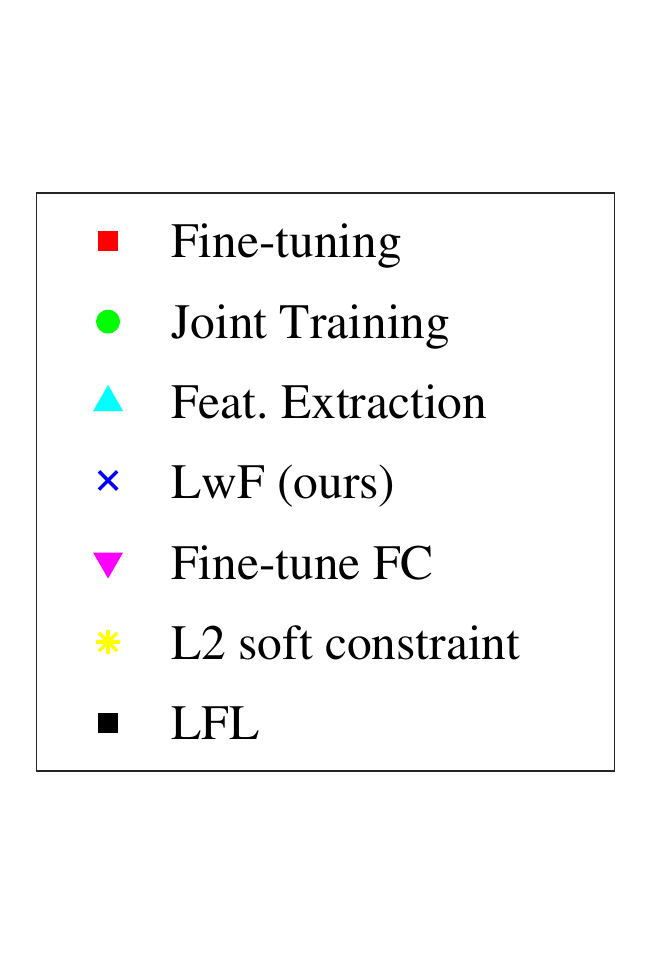}
    }
    \addtocounter{subfigure}{-1}
    \subfigure[Places365$\rightarrow$VOC]{
    \includegraphics[height=0.22\textheight]{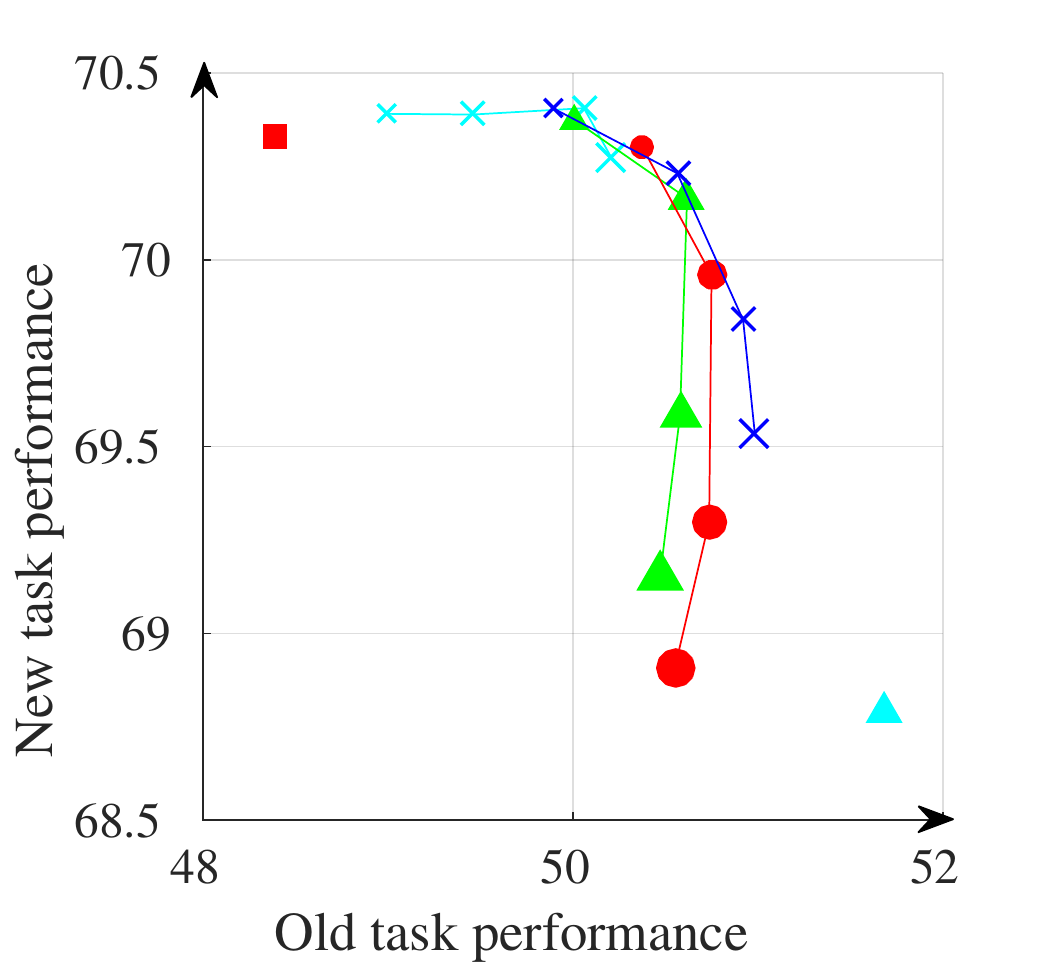}
    }
    \subfigure[ImageNet$\rightarrow$Scene]{
    \hspace{1em}
    \includegraphics[height=0.22\textheight]{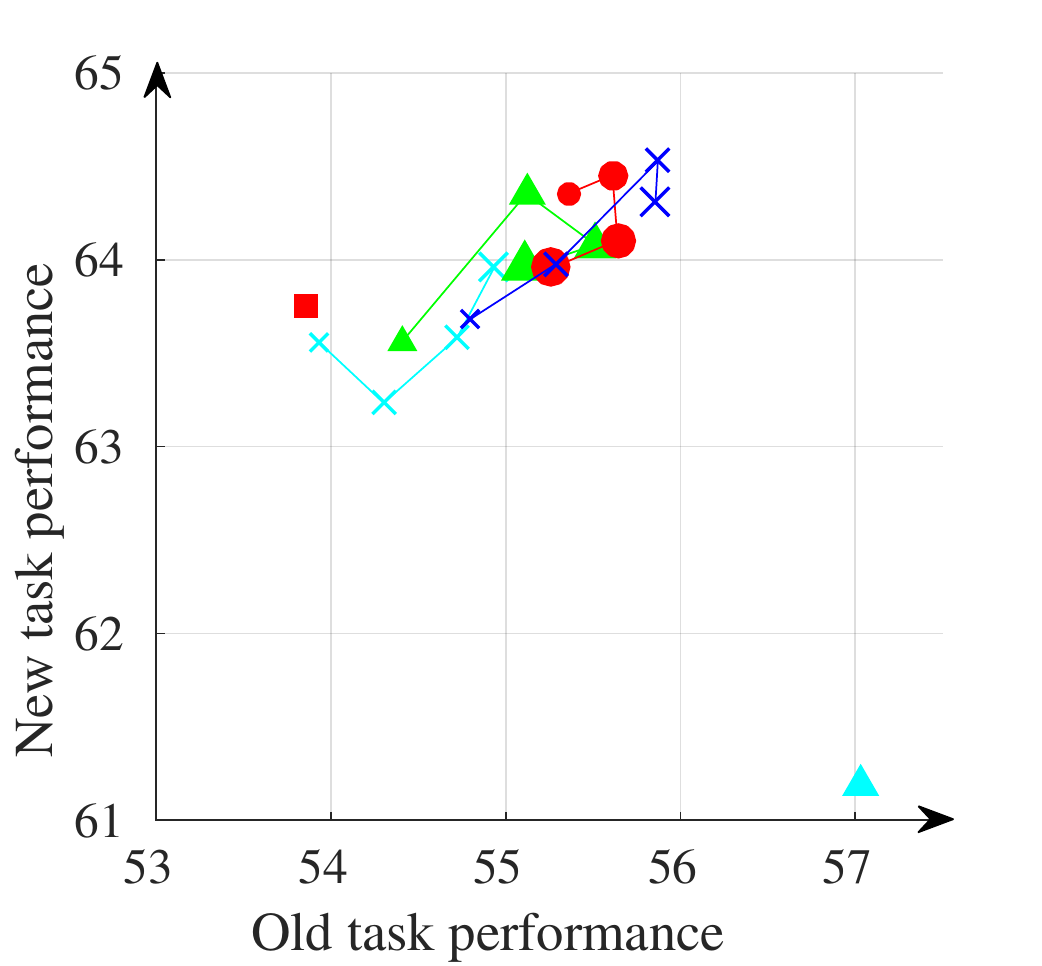}
    \hspace{1em}
    }
    \subfigure{
    \includegraphics[height=0.22\textheight]{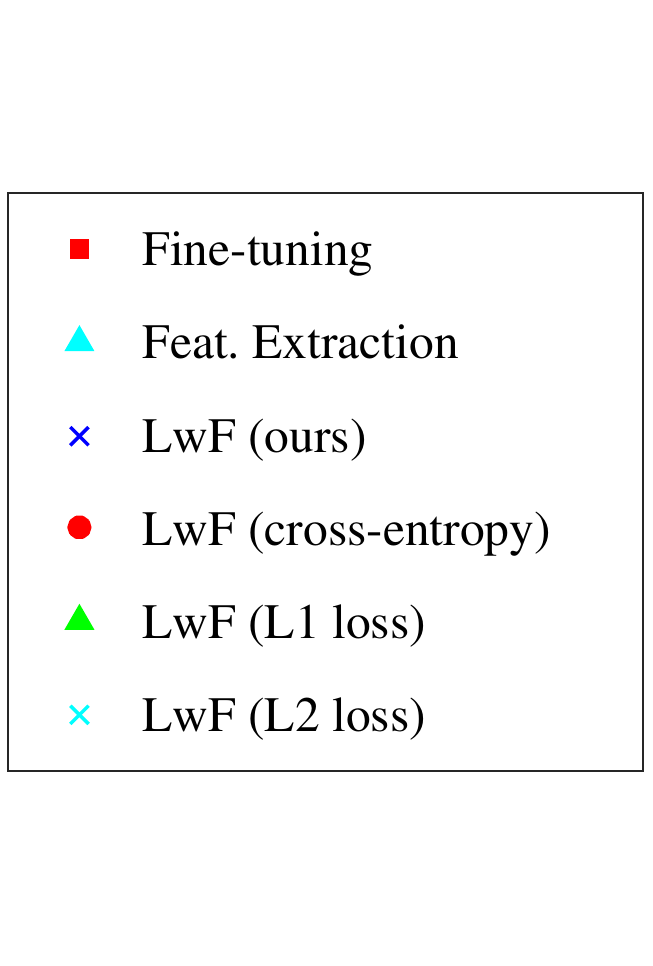}
    }
    \caption{Visualization of both new and old task performance for compared methods, some with different weights of losses. (a)(b): comparing methods; (c)(d): comparing losses. Larger symbols signifies larger $\lambda_o$, i.e. heavier weight towards response-preserving loss.}
    \label{fig:OvN}
\end{figure*}

\subsection{Design choices and alternatives}
\label{sec:alternatives}

\textbf{Choice of task-specific layers.} It is possible to regard more layers as task-specific $\theta_o$, $\theta_n$ (see Figure~\ref{fig:methods_illust_alt}(a)) instead of regarding only the output nodes as task-specific. This may provide advantage for both tasks because later layers tend to be more task specific~\cite{azizpour2014factors}. However, doing so requires more storage, as most parameters in AlexNet are in the first two fully connected layers.  Table~\ref{tab:BRANCH_EXPAND_LOWLR} shows the comparison on three task pairs. \textit{Our results do not indicate any advantage to having additional task-specific layers.}

~\\\textbf{Network expansion.} We explore another way of modifying the network structure, which we refer to as ``network expansion'', which adds nodes to some layers. This allows for extra new-task-specific information in the earlier layers while still using the original network's information.

Figure~\ref{fig:methods_illust_alt}(b) illustrates this method. We add 1024 nodes to each layer of the top 3 layers. The weights from all nodes at previous layer to the new nodes at current layer are initialized the same way Net2Net~\cite{Chen2016Net2Net} would expand a layer by copying nodes. Weights from new nodes at previous layer to the original nodes at current layer are initialized to zero. The top layer weights of the new nodes are randomly re-initialized. Then we either freeze the existing weights and fine-tune the new weights on the new task (``network expansion''), or train using Learning without Forgetting as before (``network expansion + LwF''). Note that both methods needs the network to scale quadratically with respect to the number of new tasks.

Table~\ref{tab:BRANCH_EXPAND_LOWLR} shows the comparison with our original method. \textit{Network expansion by itself performs better than feature extraction, but not as well as LwF on the new task. Network Expansion + LwF performs similarly to LwF} with additional computational cost and complexity.

~\\\textbf{Effect of lower learning rate of shared parameters.} We investigate whether simply lowering the learning rate of the shared parameters $\theta_s$ would preserve the original task performance. The result is shown in Table~\ref{tab:BRANCH_EXPAND_LOWLR}. A reduced learning rate does not prevent fine-tuning from significantly reducing original task performance, and it reduces new task performance. This shows that \textit{simply reducing the learning rate of shared layers is insufficient for original task preservation.}

~\\\textbf{L2 soft-constrained weights.}  Perhaps an obvious alternative to LwF is to keep the network parameters (instead of the response) close to the original. We compare with the baseline that adds $\frac{1}{2}\lambda_o\|w-w_0\|^2$ to the loss for fine-tuning, where $w$ and $w_0$ are flattened vectors of all shared parameters $\theta_s$ and their original values. We change the coefficient $\lambda_o$ and observe its effect on the performance. $\lambda_o$ is set to 0.15, 0.5, 1.5, 2.5 for Places365$\rightarrow$VOC, and 0.005, 0.015, 0.05, 0.15, 0.25 for ImageNet$\rightarrow$Scene.

As shown in Figure~\ref{fig:OvN}, \textit{our method outperforms this baseline, which produces a result between feature extraction (no parameter change) and fine-tuning (free parameter change)}. We believe that by regularizing the output, our method maintains old task performance better than regularizing individual parameters, since many small parameter changes could cause big changes in the outputs.

~\\\textbf{Choice of response preserving loss.} We compare the use of $L_1$, $L_2$, cross-entropy loss, and knowledge distillation loss with $T=2$ for keeping $\mathbf{y}'_o,\mathbf{\hat{y}}'_o$ similar. We test on the same task pairs as before. Figure~\ref{fig:OvN} shows our results. \textit{Results indicate our knowledge distillation loss slightly outperforms compared losses, although the advantage is not large.}

\section{Discussion}
\label{sec:discussion}

We address the problem of adapting a vision system to a new task while preserving performance on original tasks, without access to training data for the original tasks.  We propose the Learning without Forgetting method for convolutional neural networks, which can be seen as a hybrid of knowledge distillation and fine-tuning, learning parameters that are discriminative for the new task while preserving outputs for the original tasks on the training data. We show the effectiveness of our method on a number of classification tasks.

As another use-case example, we investigate using LwF in the application of tracking in Appendix~\ref{sec:MDNet}. We build on MD-Net~\cite{nam2016mdnet}, which views tracking as a template classification task. A classifier transferred from training videos is fine-tuned online to classify regions as the object or background. We propose to replace the fine-tuning step with Learning without Forgetting. We leave the details and implementation to the appendix. We observe some improvements by applying LwF, but the difference is not statistically significant.

Our work has implications for two uses.  First, if we want to expand the set of possible predictions on an existing network, our method performs similarly to joint training but is faster to train and does not require access to the training data for previous tasks.  Second, if we care only about the performance for the new task, our method often outperforms the current standard practice of fine-tuning.  Fine-tuning approaches use a low learning rate in hopes that the parameters will settle in a ``good'' local minimum not too far from the original values.  Preserving outputs on the old task is a more direct and interpretable way to to retain the important shared structures learned for the previous tasks.

We see several directions for future work.  We have demonstrated the effectiveness of LwF for image classification and one experiment on tracking, but would like to further experiment on semantic segmentation, detection, and problems outside of computer vision.  Additionally, one could explore variants of the approach, such as maintaining a set of unlabeled images to serve as representative examples for previously learned tasks.  Theoretically, it would be interesting to bound the old task performance based on preserving outputs for a sample drawn from a different distribution.  More generally, there is a need for approaches that are suitable for online learning across different tasks, especially when classes have heavy tailed distributions.

\ifCLASSOPTIONcompsoc
  \section*{Acknowledgments}
\else
  \section*{Acknowledgment}
\fi
This work is supported in part by NSF Awards 14-46765 and 10-53768 and ONR MURI N000014-16-1-2007.

\ifCLASSOPTIONcaptionsoff
  \newpage
\fi


\bibliographystyle{IEEEtran}
\bibliography{egbib}

\begin{thebibliography}{10}
\providecommand{\url}[1]{#1}
\csname url@samestyle\endcsname
\providecommand{\newblock}{\relax}
\providecommand{\bibinfo}[2]{#2}
\providecommand{\BIBentrySTDinterwordspacing}{\spaceskip=0pt\relax}
\providecommand{\BIBentryALTinterwordstretchfactor}{4}
\providecommand{\BIBentryALTinterwordspacing}{\spaceskip=\fontdimen2\font plus
\BIBentryALTinterwordstretchfactor\fontdimen3\font minus
  \fontdimen4\font\relax}
\providecommand{\BIBforeignlanguage}[2]{{%
\expandafter\ifx\csname l@#1\endcsname\relax
\typeout{** WARNING: IEEEtran.bst: No hyphenation pattern has been}%
\typeout{** loaded for the language `#1'. Using the pattern for}%
\typeout{** the default language instead.}%
\else
\language=\csname l@#1\endcsname
\fi
#2}}
\providecommand{\BIBdecl}{\relax}
\BIBdecl

\bibitem{mccloskey1989catastrophic}
M.~McCloskey and N.~J. Cohen, ``Catastrophic interference in connectionist
  networks: The sequential learning problem,'' \emph{Psychology of learning and
  motivation}, vol.~24, pp. 109--165, 1989.

\bibitem{goodfellow2013empirical}
I.~J. Goodfellow, M.~Mirza, D.~Xiao, A.~Courville, and Y.~Bengio, ``An
  empirical investigation of catastrophic forgetting in gradient-based neural
  networks,'' \emph{arXiv preprint arXiv:1312.6211}, 2013.

\bibitem{krizhevsky2012imagenet}
A.~Krizhevsky, I.~Sutskever, and G.~E. Hinton, ``Imagenet classification with
  deep convolutional neural networks,'' in \emph{Advances in neural information
  processing systems}, 2012, pp. 1097--1105.

\bibitem{ILSVRC15}
O.~Russakovsky, J.~Deng, H.~Su, J.~Krause, S.~Satheesh, S.~Ma, Z.~Huang,
  A.~Karpathy, A.~Khosla, M.~Bernstein, A.~C. Berg, and L.~Fei-Fei, ``{ImageNet
  Large Scale Visual Recognition Challenge},'' \emph{International Journal of
  Computer Vision (IJCV)}, vol. 115, no.~3, pp. 211--252, 2015.

\bibitem{Donahue_ICML2014DeCAF}
J.~Donahue, Y.~Jia, O.~Vinyals, J.~Hoffman, N.~Zhang, E.~Tzeng, and T.~Darrell,
  ``Decaf: A deep convolutional activation feature for generic visual
  recognition,'' in \emph{International Conference in Machine Learning (ICML)},
  2014.

\bibitem{Girshick_2014_CVPRRCNN}
R.~Girshick, J.~Donahue, T.~Darrell, and J.~Malik, ``Rich feature hierarchies
  for accurate object detection and semantic segmentation,'' in \emph{The IEEE
  Conference on Computer Vision and Pattern Recognition (CVPR)}, June 2014.

\bibitem{caruana1997multitask}
R.~Caruana, ``Multitask learning,'' \emph{Machine learning}, vol.~28, no.~1,
  pp. 41--75, 1997.

\bibitem{furlanello2016active}
T.~Furlanello, J.~Zhao, A.~M. Saxe, L.~Itti, and B.~S. Tjan, ``Active long term
  memory networks,'' \emph{arXiv preprint arXiv:1606.02355}, 2016.

\bibitem{jung2016less}
H.~Jung, J.~Ju, M.~Jung, and J.~Kim, ``Less-forgetting learning in deep neural
  networks,'' \emph{arXiv preprint arXiv:1607.00122}, 2016.

\bibitem{li2016learning}
Z.~Li and D.~Hoiem, ``Learning without forgetting,'' in \emph{European
  Conference on Computer Vision}.\hskip 1em plus 0.5em minus 0.4em\relax
  Springer, 2016, pp. 614--629.

\bibitem{hinton2015distilling}
G.~Hinton, O.~Vinyals, and J.~Dean, ``Distilling the knowledge in a neural
  network,'' in \emph{NIPS Workshop}, 2014.

\bibitem{razavian2014cnnfeatures}
A.~Razavian, H.~Azizpour, J.~Sullivan, and S.~Carlsson, ``Cnn features
  off-the-shelf: an astounding baseline for recognition,'' in \emph{Proceedings
  of the IEEE Conference on Computer Vision and Pattern Recognition Workshops},
  2014, pp. 806--813.

\bibitem{azizpour2014factors}
H.~Azizpour, A.~Razavian, J.~Sullivan, A.~Maki, and S.~Carlsson, ``Factors of
  transferability for a generic convnet representation,'' in \emph{IEEE
  Transactions on Pattern Analysis \& Machine Intelligence}, 2014.

\bibitem{agrawal14analyzing}
P.~Agrawal, R.~Girshick, and J.~Malik, ``Analyzing the performance of
  multilayer neural networks for object recognition,'' in \emph{Proceedings of
  the European Conference on Computer Vision ({ECCV})}, 2014.

\bibitem{yosinski2014howtransferable}
J.~Yosinski, J.~Clune, Y.~Bengio, and H.~Lipson, ``How transferable are
  features in deep neural networks?'' in \emph{Advances in Neural Information
  Processing Systems}, 2014, pp. 3320--3328.

\bibitem{chapelle2011boosted}
O.~Chapelle, P.~Shivaswamy, S.~Vadrevu, K.~Weinberger, Y.~Zhang, and B.~Tseng,
  ``Boosted multi-task learning,'' \emph{Machine learning}, vol.~85, no. 1-2,
  pp. 149--173, 2011.

\bibitem{terekhov2015knowledge}
A.~V. Terekhov, G.~Montone, and J.~K. O’Regan, ``Knowledge transfer in deep
  block-modular neural networks,'' in \emph{Biomimetic and Biohybrid
  Systems}.\hskip 1em plus 0.5em minus 0.4em\relax Springer, 2015, pp.
  268--279.

\bibitem{rusu2016progressive}
A.~A. Rusu, N.~C. Rabinowitz, G.~Desjardins, H.~Soyer, J.~Kirkpatrick,
  K.~Kavukcuoglu, R.~Pascanu, and R.~Hadsell, ``Progressive neural networks,''
  \emph{arXiv preprint arXiv:1606.04671}, 2016.

\bibitem{romero2015fitnets}
A.~Romero, N.~Ballas, S.~E. Kahou, A.~Chassang, C.~Gatta, and Y.~Bengio,
  ``Fitnets: Hints for thin deep nets,'' in \emph{Proceedings of the
  International Conference on Learning Representations (ICLR)}, 2015.

\bibitem{Chen2016Net2Net}
T.~Chen, I.~Goodfellow, and J.~Shlens, ``Net2net: Accelerating learning via
  knowledge transfer,'' in \emph{Proceedings of the International Conference on
  Learning Representations (ICLR)}, 2016, p. to appear.

\bibitem{pan2010survey}
S.~J. Pan and Q.~Yang, ``A survey on transfer learning,'' \emph{Knowledge and
  Data Engineering, IEEE Transactions on}, vol.~22, no.~10, pp. 1345--1359,
  2010.

\bibitem{long2015learning}
M.~Long and J.~Wang, ``Learning transferable features with deep adaptation
  networks,'' \emph{arXiv preprint arXiv:1502.02791}, 2015.

\bibitem{tzeng2015simultaneous}
E.~Tzeng, J.~Hoffman, T.~Darrell, and K.~Saenko, ``Simultaneous deep transfer
  across domains and tasks,'' in \emph{Proceedings of the IEEE International
  Conference on Computer Vision}, 2015, pp. 4068--4076.

\bibitem{thrun1998lifelong}
S.~Thrun, ``Lifelong learning algorithms,'' in \emph{Learning to learn}.\hskip
  1em plus 0.5em minus 0.4em\relax Springer, 1998, pp. 181--209.

\bibitem{NELL-aaai15}
T.~Mitchell, W.~Cohen, E.~Hruschka, P.~Talukdar, J.~Betteridge, A.~Carlson,
  B.~Dalvi, M.~Gardner, B.~Kisiel, J.~Krishnamurthy, N.~Lao, K.~Mazaitis,
  T.~Mohamed, N.~Nakashole, E.~Platanios, A.~Ritter, M.~Samadi, B.~Settles,
  R.~Wang, D.~Wijaya, A.~Gupta, X.~Chen, A.~Saparov, M.~Greaves, and
  J.~Welling, ``Never-ending learning,'' in \emph{Proceedings of the
  Twenty-Ninth AAAI Conference on Artificial Intelligence (AAAI-15)}, 2015.

\bibitem{eaton2013ella}
E.~Eaton and P.~L. Ruvolo, ``Ella: An efficient lifelong learning algorithm,''
  in \emph{Proceedings of the 30th International Conference on Machine
  Learning}, 2013, pp. 507--515.

\bibitem{Simonyan14cVGG}
K.~Simonyan and A.~Zisserman, ``Very deep convolutional networks for
  large-scale image recognition,'' \emph{CoRR}, vol. abs/1409.1556, 2014.

\bibitem{vedaldi15matconvnet}
A.~Vedaldi and K.~Lenc, ``Matconvnet -- convolutional neural networks for
  matlab,'' in \emph{Proceeding of the {ACM} Int. Conf. on Multimedia}, 2015.

\bibitem{glorot2010understanding}
X.~Glorot and Y.~Bengio, ``Understanding the difficulty of training deep
  feedforward neural networks.'' in \emph{Aistats}, vol.~9, 2010, pp. 249--256.

\bibitem{zhou2016places}
B.~Zhou, A.~Khosla, A.~Lapedriza, A.~Torralba, and A.~Oliva, ``Places: An image
  database for deep scene understanding,'' \emph{arXiv preprint
  arXiv:1610.02055}, 2016.

\bibitem{Everingham15VOCretrospect}
M.~Everingham, S.~M.~A. Eslami, L.~Van~Gool, C.~K.~I. Williams, J.~Winn, and
  A.~Zisserman, ``The pascal visual object classes challenge: A
  retrospective,'' \emph{International Journal of Computer Vision}, vol. 111,
  no.~1, pp. 98--136, Jan. 2015.

\bibitem{WahCUB_200_2011}
C.~Wah, S.~Branson, P.~Welinder, P.~Perona, and S.~Belongie, ``{The
  Caltech-UCSD Birds-200-2011 Dataset},'' California Institute of Technology,
  Tech. Rep. CNS-TR-2011-001, 2011.

\bibitem{quattoni2009recognizingindoor}
A.~Quattoni and A.~Torralba, ``Recognizing indoor scenes,'' in \emph{Computer
  Vision and Pattern Recognition, 2009. CVPR 2009. IEEE Conference on}, 2009,
  pp. 413--420.

\bibitem{lecun1998gradient}
Y.~LeCun, L.~Bottou, Y.~Bengio, and P.~Haffner, ``Gradient-based learning
  applied to document recognition,'' \emph{Proceedings of the IEEE}, vol.~86,
  no.~11, pp. 2278--2324, 1998.

\bibitem{nam2016mdnet}
H.~Nam and B.~Han, ``Learning multi-domain convolutional neural networks for
  visual tracking,'' in \emph{The IEEE Conference on Computer Vision and
  Pattern Recognition (CVPR)}, June 2016.

\bibitem{Kristan2015a}
M.~Kristan, J.~Matas, A.~Leonardis, M.~Felsberg, L.~\v{C}ehovin, G.~Fernandez,
  T.~Vojir, G.~H\"{a}ger, G.~Nebehay, R.~Pflugfelder, A.~Gupta, A.~Bibi,
  A.~Luke\v{z}i\v{c}, A.~Garcia-Martin, A.~Saffari, A.~Petrosino, A.~S.
  Montero, A.~Varfolomieiev, A.~Baskurt, B.~Zhao, B.~Ghanem, B.~Martinez,
  B.~Lee, B.~Han, C.~Wang, C.~Garcia, C.~Zhang, C.~Schmid, D.~Tao, D.~Kim,
  D.~Huang, D.~Prokhorov, D.~Du, D.-Y. Yeung, E.~Ribeiro, F.~S. Khan,
  F.~Porikli, F.~Bunyak, G.~Zhu, G.~Seetharaman, H.~Kieritz, H.~T. Yau, H.~Li,
  H.~Qi, H.~Bischof, H.~Possegger, H.~Lee, H.~Nam, I.~Bogun, J.~chan Jeong,
  J.~il~Cho, J.-Y. Lee, J.~Zhu, J.~Shi, J.~Li, J.~Jia, J.~Feng, J.~Gao, J.~Y.
  Choi, J.-W. Kim, J.~Lang, J.~M. Martinez, J.~Choi, J.~Xing, K.~Xue,
  K.~Palaniappan, K.~Lebeda, K.~Alahari, K.~Gao, K.~Yun, K.~H. Wong, L.~Luo,
  L.~Ma, L.~Ke, L.~Wen, L.~Bertinetto, M.~Pootschi, M.~Maresca, M.~Danelljan,
  M.~Wen, M.~Zhang, M.~Arens, M.~Valstar, M.~Tang, M.-C. Chang, M.~H. Khan,
  N.~Fan, N.~Wang, O.~Miksik, P.~H.~S. Torr, Q.~Wang, R.~Martin-Nieto,
  R.~Pelapur, R.~Bowden, R.~Laganiere, S.~Moujtahid, S.~Hare, S.~Hadfield,
  S.~Lyu, S.~Li, S.-C. Zhu, S.~Becker, S.~Duffner, S.~L. Hicks, S.~Golodetz,
  S.~Choi, T.~Wu, T.~Mauthner, T.~Pridmore, W.~Hu, W.~H\"{u}bner, X.~Wang,
  X.~Li, X.~Shi, X.~Zhao, X.~Mei, Y.~Shizeng, Y.~Hua, Y.~Li, Y.~Lu, Y.~Li,
  Z.~Chen, Z.~Huang, Z.~Chen, Z.~Zhang, and Z.~He, ``The visual object tracking
  vot2015 challenge results,'' in \emph{Visual Object Tracking Workshop 2015 at
  ICCV2015}, Dec 2015.

\bibitem{wu2015object}
Y.~Wu, J.~Lim, and M.-H. Yang, ``Object tracking benchmark,'' \emph{IEEE
  Transactions on Pattern Analysis and Machine Intelligence}, vol.~37, no.~9,
  pp. 1834--1848, 2015.

\end{thebibliography}
%
%
%


\appendices
\section{Tracking with MD-Net using LwF}
\label{sec:MDNet}
To analyze the ability of Learning without Forgetting to generalize beyond classification tasks, we examine the use-case of improving general object tracking in videos. The task is to find the bounding box of the tracked object as each image frame is given, where the very first frame's ground-truth bounding box is known. Usually the algorithm should be causal, i.e. result of frame $t$ should not depend on image frames $t+1$ and onward.

We base our method on MD-Net~\cite{nam2016mdnet}, a state-of-the-art tracker that poses tracking as a template classification task. It is unique in that it uses fine-tuning to transfer from a general network jointly trained on a number of videos to a classifier for a specific test video. Fine-tuning may potentially cause undue drift from original parameters. We hypothesize that replacing it with LwF will be more effective. In our experiment, using LwF slightly improves over MD-Net, but the difference is not statistically significant.

\subsection{MD-Net}
MD-Net tracks an object by sampling bounding boxes in the proximity of the bounding box in the last frame, and using a classifier to classify each box as the foreground object or background clutter. The algorithm picks the bounding box with the highest foreground score, apply a bounding box regression, and report the regression result. The uniqueness of MD-Net comes from the way the classifier is trained. In order to obtain a general representation of objects suitable for video tracking, MD-Net pretrains a 6-layer multi-domain neural network for classifying foreground versus background bounding boxes for 80 different sequences. The convolutional layers (\verb|conv1|-\verb|conv3|) are initialized from the VGG-M~\cite{Simonyan14cVGG} network. Data from different sequences are considered different domains, therefore the pretraining procedure is the same as joint training with the first five layers shared, and the final layer domain-specific -- thus the name ``multi-domain convolutional neural network''. In this way the topmost shared layer provides a general representation of tracked objects in videos.

At test time, all final layers are discarded, replaced by a randomly initialized layer for the test video. The convolutional layers are frozen and the rest of the network are trained on samples from the first frame. A bounding box regression layer is trained on top of the convolutional layers from the first frame's data, and is kept unchanged. Then MD-Net starts to track the object in consequent frames, occasionally training the fully-connected layers using data from previous frames sampled from hard-negative mining. We refer our readers to the original paper~\cite{nam2016mdnet} for details.

MD-Net is evaluated on, among other datasets, VOT 2015~\cite{Kristan2015a} -- a general object tracking benchmark and challenge. VOT 2015 mainly uses the expected average overlap measure (over 15 runs of a method), which is a combination of tracking accuracy and robustness, to evaluate the trackers. We refer our readers to the VOT 2015 report~\cite{Kristan2015a} for details.

\subsection{MD-Net + LwF}
The online training method used in test time can be seen as the fine-tune~FC baseline. Since our method outperforms fine-tune~FC on the new task most of the time, we experimented with using Learning without Forgetting to perform the online training step. Hopefully, the additional regularization can benefit these updates, since the new task data are from a very confined space (crops from one single video).

Specifically, we pretrained the network using code provided by the authors. At test time, instead of throwing away the task-specific final layers, we keep them as old task parameters. We also keep a copy of the original pretrained network to compute the responses of the old tasks, because the new task data are obtained online when the network will have changed. While performing online training, we run the training data on the old network to compute the responses, and use the Learning without Forgetting loss on the updated multi-task network. A loss balance of $\lambda_o=1.6$ is used. The convolutional layers are left frozen, like in MD-Net. 

The rest of the training, tracking and testing procedure is left unchanged. Like MD-Net, we pretrain using OTB-100~\cite{wu2015object}, excluding the sequences appearing in VOT 2015. Then the tracking algorithm is tested on VOT 2015 for 15 runs.

\begin{table}[t]
\centering
\caption{MD-Net compared to MD-Net + LwF on VOT 2015. Our method seems to improve upon MD-Net, but the difference is not statistically significant.}
\begin{tabular}{@{\extracolsep{4pt}} r c@{} }
  \toprule
  & Expected \\
  & Average Overlap \\
  \midrule
MD-Net~\cite{nam2016mdnet}	& 0.373 \\
MD-Net + LwF	& 0.383  \\
  \bottomrule
\end{tabular}
\label{tab:MDNet}
\end{table}

\textbf{Results.} Table~\ref{tab:MDNet} shows the performance of our method. The two methods start from the same pre-trained network (the provided pretrained network does not contain the final layers). MD-Net~\cite{nam2016mdnet} reports slightly better performance (0.386), possibly due to randomness in the pretraining step. We observe that \textit{our method slightly improves MD-Net.} However, when we compute the expected average overlap on single runs, the scores vary greatly. We observe that \textit{the improvement is not statistically significant} ($p=0.70$ for Student's $t$-test).

\section{Split of VOC and Scene}\label{sec:split}

In Section~4.1, the multiple new task experiment, we split the new tasks, VOC and Scene, into three category groups. For VOC:
\begin{itemize}
\item Transport: aeroplane, bicycle, boat, bus, car, motorbike.
\item Animals: bird, cat, cow, dog, horse, person, sheep, train.
\item Objects: bottle, chair, diningtable, pottedplant, sofa, tvmonitor.
\end{itemize}

And for Scene:
\begin{itemize}
\item Large rooms: airport\_inside, auditorium, casino, church\_inside, cloister, concert\_hall, greenhouse, grocerystore, inside\_bus, inside\_subway, library, lobby, mall, movietheater, museum, poolinside, subway, trainstation, warehouse, winecellar.
\item Medium rooms: bakery, bar, bookstore, bowling, buffet, classroom, clothingstore, computerroom, deli, fastfood\_restaurant, florist, gameroom, gym, jewelleryshop, kindergarden, laboratorywet, laundromat, locker\_room, meeting\_room, office, pantry, restaurant, shoeshop, toystore, videostore.
\item Small rooms: artstudio, bathroom, bedroom, children\_room, closet, corridor, dentaloffice, dining\_room, elevator, garage, hairsalon, hospitalroom, kitchen, livingroom, nursery, operating\_room, prisoncell, restaurant\_kitchen, stairscase, studiomusic, tv\_studio, waitingroom
\end{itemize}
This split is also used in~\cite{li2016learning}.






\end{document}